\documentclass[11pt]{article}
\usepackage[preprint]{acl}
\usepackage{times}
\usepackage{latexsym}
\usepackage[T1]{fontenc}
\usepackage[utf8]{inputenc}
\usepackage{microtype}
\usepackage{inconsolata}

\usepackage{amsmath,amsfonts,bm}

\def\eqref#1{equation~\ref{#1}}

\def\1{\bm{1}}

\DeclareMathAlphabet{\mathsfit}{\encodingdefault}{\sfdefault}{m}{sl}
\SetMathAlphabet{\mathsfit}{bold}{\encodingdefault}{\sfdefault}{bx}{n}

\usepackage{amssymb,mathtools}
\usepackage{graphicx}
\usepackage{booktabs}
\usepackage{tabularx}
\usepackage{multirow}
\usepackage{enumitem}
\usepackage{algorithm}
\usepackage{algpseudocode}
\usepackage{xcolor}
\usepackage{url}

\newcolumntype{Y}{>{\raggedright\arraybackslash}X}

\newcommand{\method}{\textsc{TAM}}

\DeclareMathOperator{\Frontier}{Frontier}

\title{From Interaction to Capability:\\
Task Advancement as State and Supervision for GUI Agents}

\author{
  Ziyang Chen\textsuperscript{1} \quad
  Qian Li\textsuperscript{2} \quad
  Chengwei Qin\textsuperscript{1} \\
  \textsuperscript{1}The Hong Kong University of Science and Technology (Guangzhou) \\
  \textsuperscript{2}Joint SDU-NTU Centre for Artificial Intelligence Research (C-FAIR), \\
  Shandong University
}

\hypersetup{
  pdftitle={From Interaction to Capability: Task Advancement as State and Supervision for GUI Agents},
  pdfauthor={Ziyang Chen, Qian Li, Chengwei Qin}
}

\begin{document}
\maketitle

\begin{abstract}
{GUI memory should preserve the results of past actions and learn reusable operations. This is difficult in long-horizon workflows, where agents must use information across applications and useful operations are interleaved with failed attempts. We introduce \textbf{Task Advancement Memory} (\method{}), which defines task advancement as verified task progress, and supports ongoing task execution and learns reusable operations. During execution, Task Advancement State represents how verified task progress changes the remaining goals, allowing the agent to build on earlier achievements across applications. After each task, advancement supervision identifies and abstracts the interactions that produced a verified local result. The learned operations preserve the relation between required inputs, expected outcomes, and verification conditions. This retains locally successful operations from both successful and failed tasks. On MobileWorld GUI-only tasks, Advancement State alone raises success from 12.0\% to 28.2\%. Full TAM achieves a 17.6\% relative gain in task success over the strongest evaluated baseline and reduces lifecycle tokens per solved task by 38.7\%. On SPA-Bench, TAM achieves state-of-the-art task success in the single-app English Level 3 setting.}
\end{abstract}

\begin{figure}[t]
    \centering
    \includegraphics[width=\linewidth]{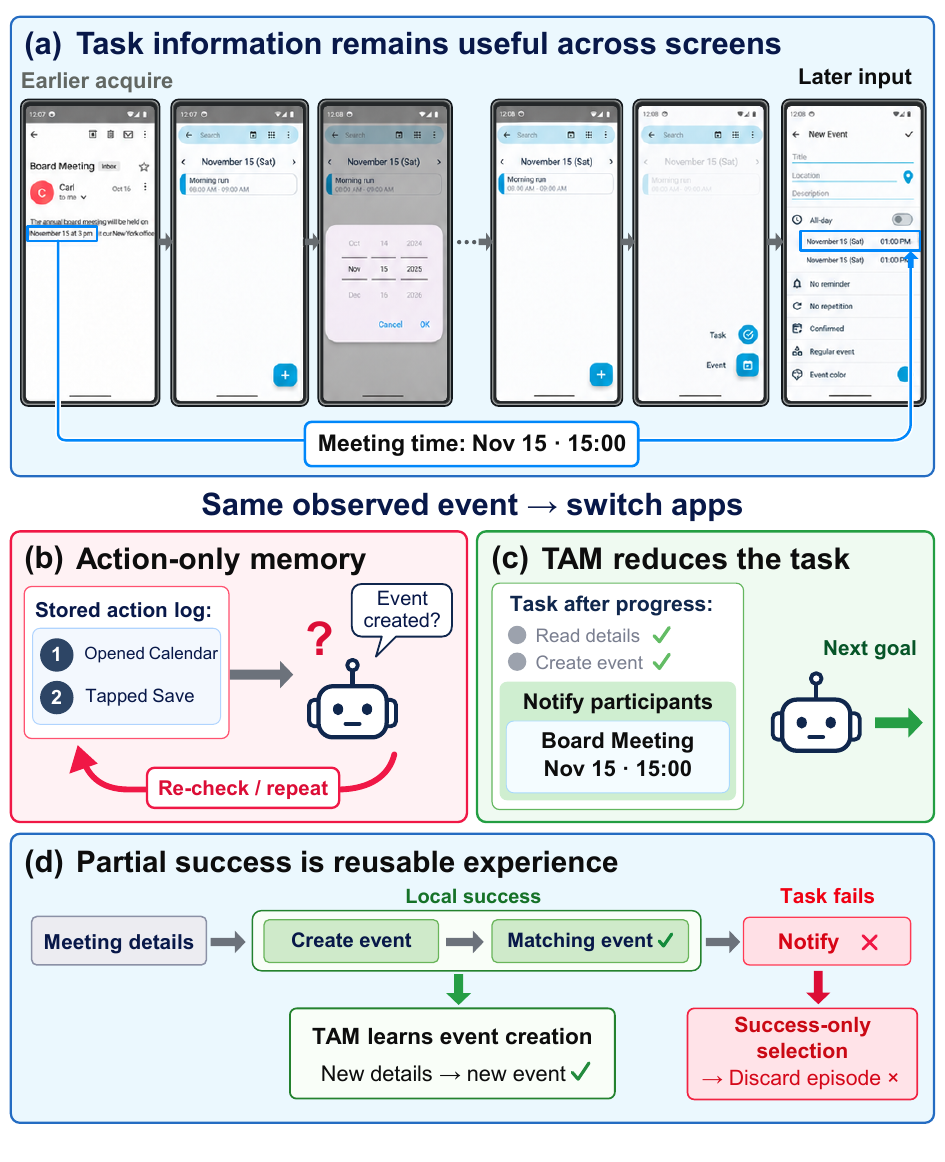}
    \caption{
        Task advancement connects execution and learning.
        (a) To create the calendar event, the agent must retain the meeting time after leaving the email.
        (b--c) Recording a save action leaves its outcome uncertain. TAM retains the confirmed event and its details, making participant notification the remaining goal.
        (d) Even if notification fails, the verified event-creation operation can be reused with new meeting details.
    }
    \label{fig:tam_motivation}
\end{figure}

\section{Introduction}
\label{sec:introduction}

GUI agents should complete workflows whose information and effects are distributed across screens and applications. Long interaction histories mix completed work with attempts and productive actions with detours, leaving the agent to reconstruct how far the task has advanced and what remains to be done. For example, after confirming a calendar event, an agent should proceed to notifying participants using the event details it has already read. Therefore, memory should represent how completed work changes the remaining task.

Recent methods improve the information available to the agent through context management, program-guided execution, and verified task state \citep{yang2025agentoccam,tian2025agentprog,liu2026atmem,zheng2026tsr,wu2026structagent,liu2026memgui}. Trajectory exemplars, induced workflows, and experience memories also transfer knowledge to later tasks \citep{zheng2024synapse,wang2025awm,wang2025mobileagente,xiao2026uimem,ouyang2025reasoningbank}. These approaches address how memory can help an agent complete remaining goals and how successful interactions can become reusable operations. However, agents still struggle to apply correctly acquired information to later goals across applications. Workflow memories tied to historical action sequences also leave the agent to adapt past routes when the interface changes. A task's final outcome also gives incomplete feedback about intermediate operations, a failed task can contain useful successes, while a successful task can include ineffective attempts.

We propose Task Advancement Memory (TAM), a memory framework that uses task progress to guide execution and learning. A \emph{task advancement}
occurs when GUI observations confirm that a previously unresolved requirement is satisfied, either by obtaining required information or by achieving a requested outcome. Each advancement reduces what the agent still needs to determine or accomplish. We call this \emph{task reduction}. For online execution, \emph{Task Advancement State} makes earlier advancements usable by later goals. It shows what information is available for each remaining goal and which goals are ready to pursue. TAM uses each advancement to supervise the selection and abstraction of productive interactions. The learned operations specify the inputs needed at entry, the result to achieve, and how to verify it. This retains useful capabilities from productive portions of both successful and failed workflows. Figure~\ref{fig:tam_motivation} illustrates these two uses of task advancement.

We evaluate \method{} on the GUI-only subset of MobileWorld
\citep{kong2026mobileworld} and the Level-3 English subset of
SPA-Bench \citep{chen2025spabench}. With
Qwen3-VL-32B-Instruct used by every listed MobileWorld method, \method{}
improves success from 29.1\% for the strongest evaluated memory baseline to 34.2\%. With GPT-4o in the SPA-Bench English Level 3 setting, it reaches 57.0\%, the highest success rate in the reported comparison. Our contributions are as follows:
\begin{itemize}[leftmargin=*,itemsep=1pt,topsep=3pt]
    \item We introduce \emph{Task Advancement State} within TAM to guide execution across applications. It represents the remaining task in terms of satisfied requirements, available inputs, and unresolved dependencies. This lets the agent build on earlier advancements when choosing its next action.
    \item We propose \emph{advancement supervision}, which converts partial task success into capabilities that can be reused in subsequent tasks. Each learned operation specifies what it needs, what it should achieve, and how to verify the outcome.
    \item We show that TAM outperforms the evaluated memory baselines in task success on MobileWorld and SPA-Bench. Cross-benchmark comparisons, component ablations, and cost analyses establish the value of task-state memory and the additional benefit of persistent operators.
\end{itemize}

\section{Related Work}
\label{sec:related_work}

\begin{figure*}[t]
    \centering
    \includegraphics[width=\textwidth]{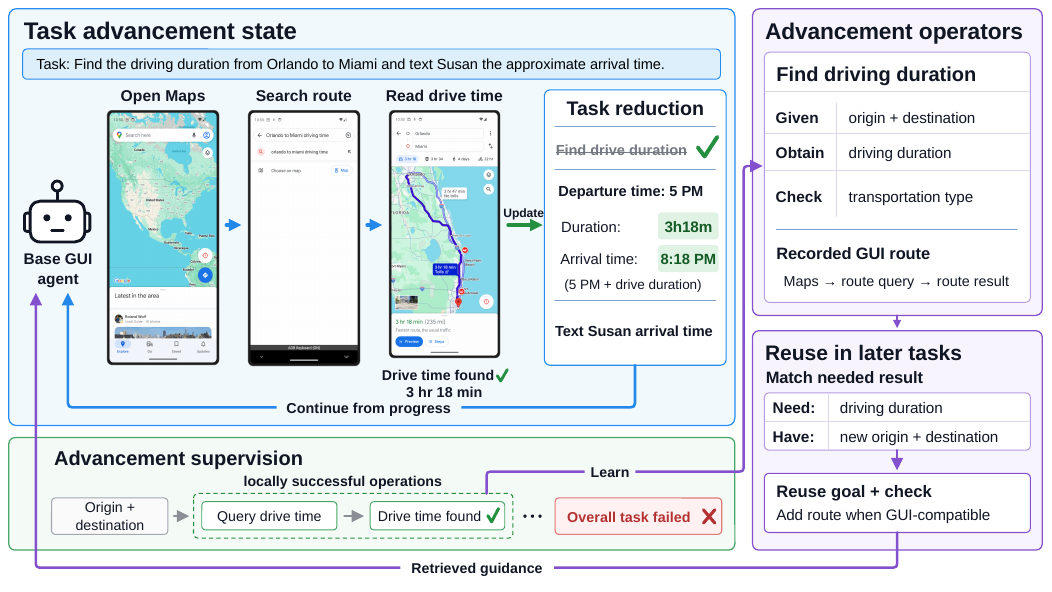}
    \caption{Overview of TAM.
    Task Advancement State guides cross-application execution through task reduction. After the route query, the remaining task is to notify Susan, with the arrival time already specified. Advancement supervision learns a reusable operator from the successful query, even if the overall task fails. Later tasks reuse the operator's goal and verification condition with new locations, adding a recorded GUI route when compatible.}
    \label{fig:tam_overview}
\end{figure*}

\paragraph{Task state and verified execution.}
Explicit task representations preserve information and unfinished work across interface changes. AgentProg represents task variables and control flow; ATMem maintains task-driving information and execution status; TSR separates durable task information from transient observations; and StructAgent combines requirements, values, evidence, and verified transitions
\citep{tian2025agentprog,liu2026atmem,zheng2026tsr,wu2026structagent}. MGA also organizes interaction around validated state changes \citep{cheng2025mga}. These methods preserve explicit task information and distinguish intended actions from observed outcomes. TAM links task results to the remaining requirements they support. These dependencies guide execution and define the inputs and outcomes of reusable operations.

\paragraph{Trajectory, workflow, and skill memory.}
Persistent experience can guide later tasks through demonstrations,
trajectories, workflows, or skills. Synapse retrieves abstracted trajectories as in-context exemplars; AWM induces workflows from offline or online experience; and ReasoningBank distills reusable reasoning strategies
\citep{zheng2024synapse,wang2025awm,ouyang2025reasoningbank}. Mobile-Agent-E stores tips and shortcuts, UI-Mem organizes workflows, subtask skills, and failure patterns, and SkillGen synthesizes and validates skills from successful and unsuccessful trajectories
\citep{wang2025mobileagente,xiao2026uimem,ma2026skillgen}. These systems reuse experience at different granularities. They adopt
different rules for locating a reusable segment, assigning local credit, and verifying the result that the segment is expected to achieve.  TAM selects interactions by the local requirement they satisfy and carries that requirement into the learned operation's expected outcome and verification condition.

\paragraph{Exploration-built interface knowledge.}
Another family constructs persistent knowledge through dedicated exploration. AppAgent builds application knowledge from exploration or demonstrations; GUI-Xplore learns action-aware structure from exploration videos; GUI-explorer generates function-aware exploration goals and mines transition knowledge; ActionEngine builds state-machine memory with an offline crawler; and EAM constructs an executable GUI graph through state-aware exploration
\citep{zhang2025appagent,sun2025guixplore,xie2025guiexplorer,zhong2026actionengine,qin2026eam}.
This approach can provide broad interface coverage before downstream tasks are attempted.

\section{Method}
\label{sec:method}

TAM uses task advancement in two ways. \emph{Task Advancement State} represents what remains to be done using the information already obtained. \emph{Advancement supervision} learns reusable operations from interactions that satisfy a task requirement. These memories guide a fixed base policy, which chooses primitive actions from the live interface (Figure~\ref{fig:tam_overview}). Appendix~\ref{app:implementation} gives the execution procedure, update rules, and learning details.

\subsection{Task Advancement as a Shared Memory Unit}
\label{sec:method_setup}

At step $t$, a base policy $\pi_\theta$ receives the instruction $g$, current GUI observation $o_t$, bounded recent history $H_t$, and TAM's bounded memory context $C_t$. It emits a primitive action in the original environment action space. TAM changes the supplied memory, without retraining the policy. A separate episode record $\mathcal T$ retains evidence for task-end review. Online state acceptance guides the current task, retrospective review evaluates which interactions should enter persistent memory.

At task initialization, an LLM-based task compiler identifies the information to acquire and the outcomes to establish from the instruction. It represents them in a graph $\mathcal Q=(V,E_Q)$, whose nodes $V$ are requirements and whose edges $E_Q$ encode their dependencies. For each requirement $q$, its dependencies $R_q$ identify the requirements that supply its inputs, and an evidence rule $V_q$ specifies how its result is recognized. An optional branch condition $\gamma_q$ specifies when the instruction requires $q$. These dependencies concern task results rather than a prescribed sequence of GUI actions. The task-local state $S_t$ records each requirement's status, known values, and supporting evidence at step $t$. Object bindings associate a value or outcome with its task entity, such as the event to which a meeting time belongs.

A task advancement records progress on a requirement: GUI observations confirm that previously missing information has been obtained or a
requested outcome has been achieved. Let $S^{-}$ and $S^{+}$ denote the task memory states before and after an evidence update. Write $z_q^{\pm}=z_{S^{\pm}}(q)$ for the status of $q$, with $\mathrm U$ and $\mathrm E$ denoting \textsc{unresolved} and \textsc{established}. An advancement is
\begin{equation}
    \operatorname{Adv}(q)
    =\mathbf 1\!\left[(z_q^{-},z_q^{+})=(\mathrm U,\mathrm E)\right]V_q(\xi),
    \label{eq:advancement}
\end{equation}
where $\xi$ is the supporting observation and
$V_q(\xi)\in\{0,1\}$ indicates whether TAM accepts it as evidence that $q$ is satisfied. Model-based extraction assesses the observed fact or effect; schema and evidence-reference checks associate it with the specified requirement. A single screen can establish several requirements. Each advancement also identifies a candidate interaction segment for later review and learning.

\begin{algorithm}[t]
\caption{Task Advancement Memory}
\label{alg:tam}
{\fontsize{10}{12}\selectfont
\begin{algorithmic}[1]
\Require instruction $g$, policy $\pi_\theta$, operator graph $\mathcal G$
\State Compile requirements $\mathcal Q$ from $g$
\State Initialize state $S$, cache $\mathcal E$, and history $H$
\State Initialize trace $\mathcal T$ and candidate segments $\mathcal B$
\While{the task is active}
    \State Observe $o$; update $S$ from supported facts
    \State Add new advancement segments to $\mathcal B$
    \State Read inputs and remaining work from $S$
    \State retrieve relevant evidence and operators for $\Frontier(S)$; assemble memory context $C$
    \State Assemble $C$ from $S$ and retrieved memory
    \State Sample $a\sim\pi_\theta(\cdot\mid g,o,H,C)$
    \State Execute $a$; observe $o'$
    \State Update $S$ from relevant effects in $o'$
    \State Add new advancement segments to $\mathcal B$
    \State Append $(o,a,o')$ to $\mathcal T$; refresh $H$
\EndWhile
\State Review $\mathcal B$ using $\mathcal T$; retain supported operations
\State Abstract entry inputs and outcomes
\State Consolidate operators into $\mathcal G$
\end{algorithmic}}
\end{algorithm}

\subsection{Advancement as State: Online Task Reduction}
\label{sec:advancement_state}

Task Advancement State turns verified progress into an explicit representation of the remaining problem. Information already obtained is bound to the requirements that need it, and satisfied requirements are removed from the remaining work. We call this \emph{task reduction}. We call the unfinished requirements with their available inputs the \emph{residual task}. Completed requirements leave this view, while their results remain in $S_t$ for later use. For example, Figure~\ref{fig:tam_overview} assumes a departure time of 5 PM. Reading a driving duration of 3 h 18 min supplies the missing value for calculating arrival at 8:18 PM. The remaining goal is to send Susan that time. TAM thus carries the result of the route query into the messaging goal across the application change.

\paragraph{Reason from the reduced task.}
Among the remaining requirements, TAM identifies those whose inputs are already available and whose branch conditions hold. This set is the \emph{Advancement Frontier}. Let $K_t$ and $U_t$ denote the established and unresolved requirements, treating an absent branch condition as true:
\begin{equation}
    F_t=\{q\in U_t:R_q\subseteq K_t,\ S_t\models\gamma_q\}.
    \label{eq:frontier}
\end{equation}
TAM uses the frontier to select evidence and operators relevant to this continuation. The actor uses the residual task and live screen to choose its next action, with earlier information already supplied to the requirements that need it.

\paragraph{Ground reductions in observed results.}
TAM updates source information from the current observation and checks relevant action outcomes using the next observation:
\begin{equation}
\begin{aligned}
    \bar S_t&=\mathcal R_{\mathrm{obs}}(S_t,o_t),\\
    S_{t+1}&=\mathcal R_{\mathrm{eff}}
    (\bar S_t,o_{t+1};\mathcal V_t).
\end{aligned}
    \label{eq:state_updates}
\end{equation}
The first update can establish visible information and object bindings. The second evaluates the set $\mathcal V_t$ of outcome requirements implicated by the action or an explicit confirmation cue. Using the observation returned by the environment, TAM distinguishes three cases: text entered into a form is a draft, clicking \texttt{Save} is an attempt, and an appropriate confirmation supports a saved result. The accepted result then becomes available to dependent requirements. This keeps progress tied to observed effects while leaving action selection with the base policy.

\paragraph{Retain visual evidence when text is insufficient.}
A task-local Evidence Cache $\mathcal E$ stores source images or crops linked to task information and its object bindings. When a requirement needs information whose source screen has left recent context, recalled evidence can restore its value or clarify which object it describes. Recalled images are identified as historical evidence. Current controls and coordinates come from the live screen.

\subsection{Advancement as Supervision: Learning Reusable Operations}
\label{sec:operator_learning}

Advancement supervision learns reusable operations from verified local results, including those within failed workflows. The challenge is both temporal and semantic: identify the productive interaction, then determine what it required and achieved. 

\paragraph{Learning from local results.}
Each advancement identifies a local result and a candidate segment of preceding interactions. When terminal evaluation nominates an outcome, it enters the same retrospective review. At task termination, an LLM-based reviewer examines the result and recorded actions and observations, including later evidence of confirmation or contradiction. It retains supported operations and selects the steps contributing to their navigation and information acquisition. Review evaluates each operation against the evidence for its local result, including when the larger task fails. TAM separates inputs needed before the retained interaction from information acquired within it. Only the former become the operation's entry requirements. For example, a route query requires an origin and destination; the driving duration is acquired by the operation itself. This makes the learned query applicable to new routes whose duration the agent still needs to find.

\paragraph{Abstract task values while preserving object identity.}
A supported segment yields an \emph{Advancement Operator}
\begin{equation}
    \kappa=\langle R_\kappa,\Delta_\kappa,V_\kappa,
    B_\kappa,\Gamma_\kappa,\mathcal A_\kappa\rangle.
    \label{eq:operator}
\end{equation}
Its components are entry requirements $R_\kappa$, expected advancement $\Delta_\kappa$, verification condition $V_\kappa$, object roles $B_\kappa$, optional applicability conditions $\Gamma_\kappa$, and reviewed GUI realizations $\mathcal A_\kappa$: recorded action sequences together with the interface conditions under which they were observed. Abstraction replaces instance-specific values with roles shared across the inputs, result, and verification target. For example, event details and the creation confirmation must refer to the same event. Concrete GUI entry conditions and action details belong to individual realizations. The persistent Advancement Operator Graph $\mathcal G$ links each
operator to its observed GUI realizations. This lets TAM reuse
the same operation while checking each route against the current interface. To consolidate a new sample, TAM compares its input--result relation with the stored operators: compatible samples refine a shared description while retaining distinct GUI realizations. Otherwise, the sample forms a separate operator.

\subsection{Operator Reuse: Task Meaning and GUI Realization}
\label{sec:operator_reuse}

An operation can remain useful across changes in its GUI implementation. TAM separates two questions: whether the operation supplies a needed result from available inputs, and whether the current interface supports a recorded route. TAM retrieves operators whose expected results and input roles match a frontier requirement. Among supported candidates, it considers both coverage of remaining requirements and the applicability and reliability of the stored GUI route. Selection retains a bounded set of alternatives across these two criteria, favoring additional requirement coverage. Appendices~\ref{app:retrieval_details} and~\ref{app:concrete_matching} specify matching, compatibility checks, and selection rules. When the required inputs are available and a stored GUI route passes the interface and reliability checks, the readout includes that route. Otherwise, a supported operator can still supply its required inputs, expected result, and verification cue, with missing inputs explicitly marked. Task knowledge therefore remains useful when a stored route needs adaptation. The actor uses this guidance to choose primitive actions from the current interface. Historical success guides the attempt; current observations determine whether the expected advancement is accepted. If no operator is relevant, ordinary execution continues with the task state and recent context.

\section{Experiments}
\label{sec:experiments}

\begin{table*}[t]
    \centering
    \small
    
    \resizebox{\linewidth}{!}{%
    \begin{tabular}{lcccc}
        \toprule
        Method & Success (\%) $\uparrow$ & Steps $\downarrow$
        & EQA@50 (\%) $\uparrow$ & LTS (M) $\downarrow$ \\
        \midrule
        Qwen3-VL-32B-Instruct \citep{bai2025qwen3vl}
            & 12.0 & -- & -- & -- \\
        AWM \citep{wang2025awm}
            & 13.7 & 25.3 & 11.9 & \textbf{0.720} \\
        Mobile-Agent-E \citep{wang2025mobileagente}
            & 15.4 & 22.9 & 12.2 & 4.472 \\
        StructAgent \citep{wu2026structagent}
            & 24.8 & 27.8 & 19.1 & $\geq 1.115^{\ast}$ \\
        GUI-explorer \citep{xie2025guiexplorer}
            & 25.6 & 27.4 & 20.6 & $\geq 1.478^{\ast}$ \\
        MemGUI-Agent (ConAct) \citep{liu2026memgui}
            & \underline{29.1} & \underline{22.4} & \underline{23.7} & 1.257 \\
        \textbf{\method{} (ours)}
            & \textbf{34.2} & \textbf{20.6} & \textbf{29.2}
            & \underline{0.771} \\
        \bottomrule
    \end{tabular}%
    }
    \vspace{2pt}

    \begin{minipage}{\linewidth}
    \footnotesize
    $^{\ast}$The archived StructAgent and GUI-explorer accounting omits autonomous exploration and knowledge-base construction, so their lifecycle cost is a lower bound.
    \end{minipage}
\caption{Results on MobileWorld GUI-only tasks. All methods use
    Qwen3-VL-32B-Instruct and a limit of 50 controller rounds. EQA@50 jointly rewards
    success and earlier completion. LTS measures the total model-token expenditure required to obtain a solution.}
    \label{tab:mobileworld}
\end{table*}

\begin{table*}[t]
    \centering
    \small
    
    \begin{tabular*}{\linewidth}{@{\extracolsep{\fill}}lccc@{}}
        \toprule
        Method & Input & Base Model & Success (\%) $\uparrow$ \\
        \midrule
        CogAgent \citep{hong2024cogagent} & screen & CogAgent & 0.0 \\
        DigiRL \citep{bai2024digirl} & screen & DigiRL & 0.0 \\
        AutoDroid \citep{wen2024autodroid} & a11y tree & GPT-4o & 12.0 \\
        SeeAct \citep{zheng2024seeact} & SoM & GPT-4o & 12.0 \\
        AppAgent \citep{zhang2025appagent} & SoM & GPT-4o & 14.0 \\
        Mobile-Agent-v2 \citep{wang2024mobileagentv2} & SoM & GPT-4o & 20.0 \\
        M3A \citep{rawles2025androidworld} & SoM & GPT-4o & 42.0 \\
        MemGUI-Agent (ConAct) \citep{liu2026memgui} & SoM & GPT-4o & 46.0 \\
        GUI-explorer \citep{xie2025guiexplorer} & SoM & GPT-4o & 53.7 \\
        \textbf{\method{} (ours)} & \textbf{SoM} & \textbf{GPT-4o} & \textbf{57.0} \\
        \bottomrule
    \end{tabular*}
\caption{Task success on SPA-Bench single-app English Level 3. Results for the first seven baselines and GUI-explorer are taken from \citet{xie2025guiexplorer}.}
    \label{tab:spabench}
\end{table*}

We evaluate task completion on two benchmarks and lifecycle cost on MobileWorld. Component ablations measure the contribution of online state and persistent operators. We also examine whether earlier results remain usable and whether local review produces reliable learning material.

\subsection{Experimental Setup}
\label{sec:experimental_setup}

\paragraph{Benchmarks.}
\textbf{MobileWorld GUI-only} tests long-horizon tasks across a broad application suite, including workflows that span multiple apps \citep{kong2026mobileworld}. We use the GUI-only interaction setting with standard mobile actions. This setting stresses the preservation of information and results across application boundaries. \textbf{SPA-Bench English Level 3} evaluates multi-stage workflows within individual smartphone applications at the benchmark's highest difficulty tier \citep{chen2025spabench}. Following GUI-explorer, we study the single-app English Level~3 setting with Set-of-Mark (SoM) observations and GPT-4o \citep{xie2025guiexplorer,openai2024gpt4o}. It complements MobileWorld by testing sustained interaction with real-world applications without requiring cross-app information transfer.

\paragraph{Backbones and memory protocol.}
Every MobileWorld method in Table~\ref{tab:mobileworld} uses Qwen3-VL-32B-Instruct and the same limit of 50 controller rounds \citep{bai2025qwen3vl}. TAM builds persistent memory from interactions in the evaluation task sequence, without a separate application-exploration stage. Task Advancement State and Evidence Cache reset between tasks, reviewed operators remain available to subsequent tasks. Component ablations preserve the base policy, action interface, and interaction budget. On SPA-Bench, we use a task-specific budget of $2g_i+1$ rounds, where $g_i$ is the reference human action count. The official evaluator scores the recorded trajectory after termination; it supplies no intermediate success feedback to the policy. Appendix~\ref{app:spabench_protocol} specifies execution, memory, and the aggregation protocol.

\paragraph{Metrics.}
\emph{Success} is acceptance by the benchmark's native evaluator and is reported for both benchmarks. We additionally report the following execution and cost metrics on MobileWorld. \emph{Steps} averages controller rounds over all runs, including failures, and is interpreted jointly with success. A controller round is one externally counted action decision. \emph{Efficiency--Quality Area at 50 rounds} (EQA@50) jointly rewards successful completion and fewer controller rounds, assigning zero credit to failures \citep{wang2026mmbenchgui}. \emph{Lifecycle Tokens per Solved Task} (LTS) divides total model-token expenditure by solved tasks, including memory construction, execution, and failed attempts.

\paragraph{Baselines.}
Alongside the memory-free backbone, we compare workflow induction with AWM, tips and shortcuts with Mobile-Agent-E, exploration-built knowledge with GUI-explorer, verified task-state and procedural memory with StructAgent, and proactive context management with MemGUI-Agent (ConAct) \citep{wang2025awm,wang2025mobileagente,xie2025guiexplorer,wu2026structagent,liu2026memgui}. These comparisons cover both state-oriented and experience-oriented approaches. Table~\ref{tab:spabench} combines our SPA-Bench evaluation of TAM and ConAct with published reference results from \citet{xie2025guiexplorer}, retaining the reported observation and backbone for each method.

\subsection{Main Results}
\label{sec:main_results}

On MobileWorld, TAM achieves 34.2\% success, improving on the strongest evaluated baseline by 5.1 percentage points (Table~\ref{tab:mobileworld}). Relative to the memory-free backbone, the gain is 22.2 points. Across all three app-count groups, TAM outperforms all four compared
baselines (Table~\ref{tab:app-count}). Its advantage over ConAct is larger on the 62 multi-app tasks, where success reaches 32.3\% versus 24.2\% (+8.1 points), than on single-app tasks (+1.8 points). The larger multi-app gain supports TAM's focus on sustaining execution across application boundaries. TAM also attains the highest EQA@50, 29.2\%, and uses 20.6 controller rounds on average, showing that the advantage also holds when successful completion is rewarded for taking fewer rounds. TAM uses 0.771M lifecycle tokens per solved task, 38.7\% below ConAct's 1.257M. AWM is slightly cheaper at 0.720M, but completes 13.7\% of tasks versus TAM's 34.2\%. TAM offers the strongest completion rate in this comparison at a lifecycle cost close to the cheapest evaluated memory method. This supports the practical value of organizing memory around established task results rather than increasing the interaction budget. On SPA-Bench English Level 3, TAM reaches 57.0\% success, 3.3 percentage points above GUI-explorer's reported 53.7\% (Table~\ref{tab:spabench}). This is the highest result in the reported comparison. Together, the two benchmarks show that TAM's benefits occur both in cross-app workflows with an open-source backbone and in difficult single-app tasks with GPT-4o.

\subsection{Component Ablations}
\label{sec:ablation}

\begin{table*}[t]
\centering
\captionsetup{position=bottom,justification=raggedright,singlelinecheck=true}
\newcommand{\metrichead}[1]{{\fontsize{8}{10.5}\selectfont#1}}
\begin{minipage}[t]{0.650\textwidth}
  \vspace{0pt}
  \centering
  \fontsize{9}{10.5}\selectfont
  \setlength{\tabcolsep}{1pt}
  \renewcommand{\arraystretch}{1.00}
  \begin{tabular*}{\linewidth}{@{\extracolsep{\fill}}lc@{\hspace{6pt}}c@{\hspace{4pt}}c@{\hspace{6pt}}c@{\hspace{4pt}}c@{\hspace{6pt}}c@{}}
    \toprule
    \multirow{2}{*}{Method}
      & \multicolumn{2}{c}{1 app (55 tasks)}
      & \multicolumn{2}{c}{2 apps (48 tasks)}
      & \multicolumn{2}{c}{3 apps (14 tasks)} \\
    \cmidrule(lr){2-3}\cmidrule(lr){4-5}\cmidrule(l){6-7}
      & \metrichead{Solved} & \metrichead{TAM Gain (pp)}
      & \metrichead{Solved} & \metrichead{TAM Gain (pp)}
      & \metrichead{Solved} & \metrichead{TAM Gain (pp)} \\
    \midrule
    ConAct       & 19 & $+1.8$  & 14 & $+6.3$  & 1 & $+14.3$ \\
    GUI-explorer & 16 & $+7.3$  & 13 & $+8.3$  & 1 & $+14.3$ \\
    AWM          & 13 & $+12.7$ &  3 & $+29.2$ & 0 & $+21.4$ \\
    StructAgent  & 17 & $+5.5$  & 11 & $+12.5$ & 1 & $+14.3$ \\
    \addlinespace[1pt]
    \textbf{\method{} (ours)} & \textbf{20} & -- & \textbf{17} & -- & \textbf{3} & -- \\
    \bottomrule
  \end{tabular*}
  \captionof{table}{Completion by app count.
  Solved: completed tasks. TAM Gain: \method{}'s absolute success-rate
  advantage over the method in that row (percentage points).
  All results use the MobileWorld GUI-only split.}
  \label{tab:app-count}
\end{minipage}\hfill
\begin{minipage}[t]{0.329\textwidth}
  \vspace{0pt}
  \centering
  \fontsize{9}{10.5}\selectfont
  \setlength{\tabcolsep}{1pt}
  \renewcommand{\arraystretch}{1.2700}
  \begin{tabular*}{\linewidth}{@{\extracolsep{\fill}}lrr@{}}
    \toprule
    \metrichead{Configuration} & \metrichead{SR (\%) $\uparrow$} & \metrichead{Steps $\downarrow$} \\
    \midrule
    Base agent    & 12.0 & -- \\
    Advancement State & 28.2 & 23.4 \\
    State + Evidence Cache & 29.1 & 25.6 \\
    State + Operator Graph & 32.5 & 20.9 \\
    \textbf{Full \method{}} & \textbf{34.2} & \textbf{20.6} \\
    \bottomrule
  \end{tabular*}
  \captionof{table}{Memory-component ablation on MobileWorld GUI-only
  tasks with Qwen3-VL-32B-Instruct and a fixed base policy.}
  \label{tab:ablation}
\end{minipage}
\end{table*}

Table~\ref{tab:ablation} shows that Task Advancement State provides the largest gain, raising success from 12.0\% to 28.2\%. This gain establishes the value of advancement state for online execution before persistent experience is added. Adding the Advancement Operator Graph to State yields 32.5\%, while average steps fall from 23.4 to 20.9. With the Evidence Cache enabled, the Operator Graph adds 5.1 points (29.1\% to 34.2\%). The Operator Graph improves completion in both cache settings. Online state provides the main gain by making earlier results available to unfinished work, persistent operators add experience about how to obtain the next needed result.

\subsection{Mechanism Analysis}
\label{sec:mechanism_analysis}

\begin{figure*}[t]
    \centering
    \includegraphics[width=\textwidth]{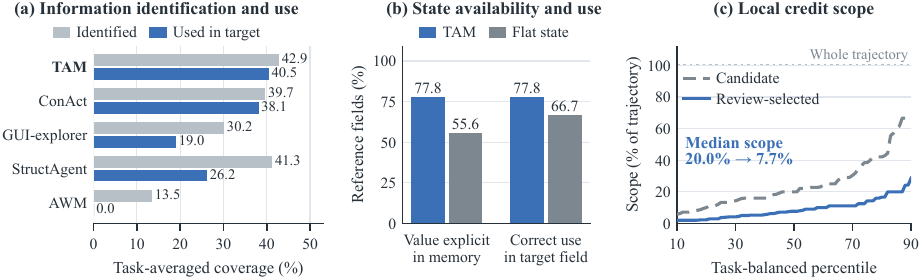}
    \caption{Mechanism analysis. (a) Task-averaged information identification and target use over 52 values from 21 tasks. (b) State availability and target use on 18 field instances with source inputs observed in both conditions. (c) Candidate and retained step counts, normalized by source-trajectory length.}
    \label{fig:tam_mechanism}
\end{figure*}

\paragraph{Identifying and using task information.}
To examine whether acquired information remains usable, we audit archived runs of tasks that
require obtaining a value from the environment and using it in a later
operation. The comparison covers the same 52 required values across 21 tasks
for all five methods. Within each task, \emph{identification coverage} is the
fraction of required values explicitly and correctly stated in online
outputs; \emph{target-use coverage} is the fraction correctly used in their
intended fields or objects. Averaging these rates equally across tasks, TAM
achieves the highest observed coverage for both identification and
target use (Figure~\ref{fig:tam_mechanism}(a)).

\paragraph{State availability and use.}
We use a flat-state control to test the value of relating stored facts to
pending requirements. The control retains fact storage and open requirements
while removing prerequisite links and their action guidance.
To compare the use of available information, we include all target fields
in the paired runs whose required source inputs were visible to both agents.
The nine eligible fields from five tasks are each assessed under two paired implementation revisions, giving 18 field instances. Over these instances,
\emph{state availability} measures
the fraction with the correct value explicitly recorded in the actor's
state. TAM achieves 77.8\% versus 55.6\% for flat state, while correct target
use reaches 77.8\% versus 66.7\%
(Figure~\ref{fig:tam_mechanism}(b)).

\paragraph{Localizing supervision for operator learning.}
For supported advancements written to operator memory, we compare candidate
segments with the steps retained after retrospective review.
Figure~\ref{fig:tam_mechanism}(c) shows the 10th--90th percentile profiles,
normalized by source-trajectory length. Review reduces the median retained segment length
from 20\% to 7.7\% of the source trajectory. Therefore, the retained supervision is concentrated on local operations rather than the full interaction history.

\subsection{Further Analyses}
Beyond the main benchmark comparisons, the appendix provides complementary
evidence on information retention, state organization, local learning
signals, experience reuse, and resource requirements.
Appendix~\ref{app:workflow_demands} classifies all 117 MobileWorld GUI-only
tasks and identifies 52 that require transferring acquired values to later
operations, establishing the scope of our field-level diagnostics.
Appendix~\ref{app:value_persistence} shows that 95.1\% of tracked values
remain in state after ten decisions, compared with 57.6\% in current XML
or recent text, demonstrating sustained access beyond the immediate
context. Appendix~\ref{app:state_control} finds that TAM assembles all
audited inputs before entry in 80\% of operations, versus 40\% for flat
state, when both agents have encountered the required source information.
This supports organizing acquired facts around pending operations.
In matched tests of access to captured facts, Appendix~\ref{app:readout}
reports 96.2\% readout accuracy versus 80.8\% for ordinary notes at similar
input size. TAM also uses 37.3\% fewer input tokens per readout than full
textual history, combining accurate access with compact context.
Appendix~\ref{app:local_learning_quality} shows that local selection
retains 88.3\% of correct material versus 64.3\% for success-only filtering,
with 93.8\% correctness among retained values, expanding the pool of
reliable learning material. Appendix~\ref{app:failed_workflows} shows
that 74.8\% of audited failed executions containing verified local results
contribute retained learning material. Recorded traces connect this
material to guidance for later tasks, demonstrating the learning value
of partial progress. Appendix~\ref{app:new_inputs} reports 83.3\%
correctness when binding changed task values and 79.3\% correct destination
use in the respective field audits, supporting the adaptation of stored
experience to new tasks. Finally, Appendix~\ref{app:computation} documents
a compact resource footprint across the archived runs, with median shares
of 5.7\% of actor text for state and guidance and 1.7\% of recorded tokens
for additional calls.

\section{Conclusion}
\label{sec:conclusion}

Task Advancement Memory makes established results useful at two timescales. During execution, \emph{task reduction} supplies acquired information to pending requirements and removes completed requirements. During learning, the interactions yield reusable operations whose entry inputs, outcome, and verification target remain aligned after abstraction. Across MobileWorld and SPA-Bench, TAM improves task completion, with ablations establishing the independent value of Advancement State and further gains from persistent operators. These findings support a task-centered foundation for GUI memory: accumulated progress makes the current problem more concrete, while productive interaction builds capabilities for future workflows.

\section*{Limitations}
\label{sec:limitations}

TAM builds task state and reusable operators from observed results. The accuracy of this process depends on model-based interpretation of task requirements and GUI evidence, especially when outcome cues are ambiguous. Improving visual grounding and outcome verification can strengthen both online execution and experience learning.

Our evaluation focuses on MobileWorld GUI-only tasks and the English Level 3 subset of SPA-Bench, covering cross-application workflows and difficult single-application tasks on mobile devices. Extending this evaluation to web and desktop workflows would assess transfer across interface types.

TAM's persistent operators accumulate across the benchmark task sequences studied here. Longer task streams and repeated application updates introduce further requirements for maintaining relevant experience. The separation between an operation's intended result and its GUI realizations provides a basis for studying adaptation as interfaces change. Future work can examine operator revision and retirement, together with their effects on retrieval quality and memory-maintenance cost.

\bibliography{iclr2027_conference}

\appendix
\section{Method and Implementation Details}
\label{app:implementation}

This appendix describes how TAM connects the requirements of a task to
observed progress and reusable operations. We follow the execution order in
Algorithm~\ref{alg:tam}: compile the requirements, maintain task advancement
state, retrieve relevant evidence and operators, and review supported local
operations for reuse. The same requirement and object identifiers connect
these stages.

\subsection{Requirements and memory records}
\label{app:record_schemas}

A requirement in $\mathcal Q$ describes either information to acquire or an
outcome to achieve. Its record contains a relation name, value type, object
roles, prerequisite requirements $R_q$, and a verification condition $V_q$.
Structured values also specify their required fields. An optional condition
$\gamma_q$ identifies the branch in which the requirement applies. For
example, an event record can contain its title, date, and time, while a
notification requirement depends on that record and the recipient's contact
information. Dependencies describe the information consumed by an operation;
the GUI route is chosen during execution.

The state $S_t$ associates each requirement with its status, current bindings,
and evidence references, as described in Section~\ref{sec:method_setup}. Bindings preserve
the relation between a value and its object: an event time belongs to the
specified event, and an address belongs to the specified recipient. Evidence
references identify the observation step and saved source artifact. The
implementation also records artifact digests so that a reused image retains
its original identity.

The Evidence Cache $\mathcal E$ stores visual support for task-local facts.
The persistent graph $\mathcal G$ stores operators and their GUI
realizations. An operator records its entry inputs, expected advancement,
verification condition, object roles, and applicability conditions. Its
realizations retain the corresponding action sequences and interface
contexts. This separates the result an operation provides from the route
used to obtain it. Task state and cached evidence reset at the task boundary;
reviewed operators remain available to subsequent tasks.

\subsection{Compiling the task requirements}
\label{app:goal_compilation}

At task start, the LLM-based compiler receives the natural-language instruction and
proposes information and outcome requirements in JSON. It assigns types, names
the necessary fields, and specifies which requirements supply inputs to each
outcome. For a conditional request, it also represents the condition and its
dependent requirement. Compilation operates on the instruction; it does not
require advance access to the applications or their task-specific contents.

The implementation combines the proposal with schema normalization. This
resolves requirement references, checks field types, and preserves prerequisite
relations when bounding the graph size. The default target is 12 requirements;
dependencies needed by a retained requirement are preserved. Explicit
instruction values, such as a requested recipient, subject, or message, are
kept as constraints on the corresponding outcome. They specify what the agent
must achieve and are kept separate from observations of what it has achieved.

Consider the request to read an event's details and notify a contact. The
compiler separates the event information, the contact endpoint, and the
notification outcome. If the task also asks for an independent numerical
answer, that calculation is connected only to the outputs that use it. The
resulting graph therefore preserves independent branches of the request
instead of requiring every intermediate result before any outcome can be
pursued.

\subsection{Updating and reading task advancement state}
\label{app:evidence_reduction}

The actor reports task-relevant observations alongside its action. Each
observation update identifies a requirement, a value or structured fields,
and the evidence that supports it. The update format distinguishes source
information, derived values, visible drafts, collection coverage, and
observed effects. These roles retain the difference between reading an
event's time, entering that time into a form, and observing a saved event.
The runtime associates the proposed update with its requirement and source
observation, checks the field structure, and updates the corresponding
binding. Model-based extraction interprets the observed content; reference
and schema checks preserve its association with the task and artifact.

Source information can be established from the current observation. Relevant
outcomes are assessed from the observation returned after an action, using
the requirement's verification condition. For instance, text visible in a
message composer describes a draft; the sent-message view provides evidence
about the resulting message. Both the content and its recipient matter when
the task specifies them. The two updates in Eq.~\ref{eq:state_updates} thus
use the observations already produced by the interaction loop.

Structured requirements retain missing fields until they are supplied.
Collections preserve observed members separately from evidence that the
requested scope has been covered. Similarly, an absence claim refers to a
particular source and search scope. Contradictory observations leave the
affected requirement conflicted until further evidence resolves it. This
keeps later inputs tied to the information that has been established.

\paragraph{Presenting the remaining work.}
The actor receives the established bindings and a requirement readout that
distinguishes available inputs from missing inputs. An input shared by
several requirements is referenced from the same state record. This is the implementation of the residual-task representation described in
Section~\ref{sec:advancement_state}: completed requirements no longer constitute
pending work, while their values remain available to their consumers.
The frontier $F_t$ identifies unresolved requirements whose dependencies and
branch conditions are satisfied.

In the event example, reading the event details supplies the notification's
content. After the contact endpoint is established, the notification has its
required inputs. The actor can proceed using those values even after leaving
the source applications. The readout organizes the next decision; it does not
issue an action or replace the original task instruction. TAM's full-method
configuration leaves primitive action selection to the base policy.

\subsection{Recalling visual evidence}
\label{app:evidence_cache}

A cache entry links a saved screenshot or crop to the accepted relation,
object binding, and source step. The cache is queried when a current
requirement needs information whose source is no longer represented in the
actor's recent visual context. Selection follows the requirement's input
dependencies and matches the corresponding bindings. If several facts come
from the same image, that image is attached once.

The implementation forms a packet covering the required source evidence
within a limit of three recalled images. Evidence already present in the
actor context contributes to that coverage. Each recalled image is marked
as a past observation and supplied with a reference identifier. When the
actor reads a value from it, the observation update cites that identifier.
The resulting value follows the ordinary state-update path. Current control
locations and coordinates are grounded in the live observation.

Cache selection does not invoke another model. The selected images enter
the ordinary actor call, where their visual tokens are included in the cost.
If no matching packet is available, execution proceeds with the current
observation, recent history, and task state.

\subsection{Reviewing local operations and abstracting their inputs}
\label{app:retrospective_review}

An accepted advancement records the affected requirement, its bindings,
supporting evidence, and the preceding interaction segment. This initial
segment retains context before the use of relevant inputs. It is a candidate
for review: some of its interactions may be navigation, repeated attempts,
or work unrelated to the established result.

At task end, the reviewer receives the candidates, their recorded actions,
the result observations, and available later evidence. It decides whether
each local result is supported, selects the steps to retain, and cites the
supporting artifacts. Later observations can confirm the result or reveal
a reversal or object mismatch. Terminal success can nominate an outcome candidate for review, it does not replace the local evidence judgment. The independent outcome-quality audit excludes such evaluator-nominated candidates (Appendix~\ref{app:local_learning_quality}). Correctly reading an event remains a useful operation
when a later notification fails; a successful task can likewise contain
unproductive attempts that need not enter its reusable operation.

\paragraph{Identifying the entry inputs.}
TAM saves the accepted task facts available before each step. Once the
reviewer selects a segment, the snapshot at its retained start identifies
which inputs were already available. Changes within the segment identify
information acquired by the operation itself. This distinction determines
$R_\kappa$: a value first obtained inside the retained segment is an internal
result, rather than an input required before the operation starts. If review
chooses a later start, the corresponding later snapshot supplies its entry
state. The implementation preserves the navigation and information-acquisition
steps needed to support the selected local result.

For example, a segment that opens a calendar and reads an event can learn
an operation for acquiring event details. A subsequent segment that starts
with those details and sends a notification can learn an operation that
consumes them. The second operation's inputs should not be assigned to the
first merely because both appear in the same completed workflow. Conversely,
reviewing a segment from a later starting point must use the information
available at that point, rather than the task's final state.

\paragraph{Consolidating an operation.}
For a supported segment, the reviewer provides a short description of the
entry inputs, local result, and reusable method. It replaces task-specific
values with object roles and compares the input--result relation with nearby
stored operators. Compatible samples refine one operator while preserving
their individual GUI realizations. Materially different inputs or outcomes
produce separate operators. Different names, dates, or layouts alone do not
require a separate semantic operation.

For the route query in Figure~\ref{fig:tam_overview}, the entry snapshot contains Orlando and Miami as the origin and destination. The driving duration is absent at entry and is obtained from the route-result screen. Abstraction replaces the two locations with origin and destination roles. The operator requires those roles, returns their driving duration, and checks that the displayed route uses the requested transportation type. Its recorded realization is Maps $\rightarrow$ route query $\rightarrow$ route result. The duration remains the result to obtain in a new task, rather than a prerequisite copied from the completed source trajectory.

The stored record links this description to the retained steps and evidence.
The expected result and verification condition refer to the same object roles
as the inputs. The graph can therefore preserve an operation such as creating
an event from given details while retaining different observed routes through
the calendar interface. New evidence refines the shared description while
preserving each observed route.

\subsection{Retrieval and actor readout}
\label{app:retrieval_details}

Retrieval begins with the remaining requirements and their available inputs.
TAM compares candidate operators with the needed result and checks whether
their input roles and applicability conditions fit the current task. It then
assesses each stored realization against the current application and
interface context. These are the task-relevance and GUI-compatibility
questions introduced in Section~\ref{sec:operator_reuse}.

The readout includes the operation's required inputs, expected advancement,
and verification cue. When a realization fits the live interface, it also
includes the corresponding route. Otherwise, the semantic operation still
describes a useful result and how to recognize it, while the actor chooses
a route from the current screen. Candidate selection retains a bounded set
of relevant operations and avoids filling the context with equivalent
routes. The retrieved-memory budget applies to this operator readout;
task state and recent observations have their own context limits.

Applying an operator uses the current task's bindings. Historical task values
remain part of the source record, while the actor grounds the present action
on the live interface. The expected advancement becomes an established
requirement only when current observations support it. The same state-update
process therefore connects learning from one task with verification in the
next.

\subsection{Model interfaces and call accounting}
\label{app:implementation_limits}

Table~\ref{tab:app_model_interfaces} summarizes the inputs and outputs of
the implementation's model calls. Requirements, observation references, and retained
step identifiers provide the common structure across calls. The actor also
maintains a concise execution summary containing findings, attempted routes,
and remaining work alongside the structured task state.

\begin{table}[t]
\centering
\small
\setlength{\tabcolsep}{3pt}
\begin{tabularx}{\columnwidth}{@{}p{0.18\columnwidth}YY@{}}
\toprule
Call & Inputs & Outputs \\
\midrule
Compiler & Instruction and requirement schema & Requirements, types, dependencies, and conditions \\
Actor & Instruction, live observation, recent context, and TAM readout & Observation updates, execution summary, and one primitive action \\
Outcome extraction & Relevant unresolved outcome and observed action result & Proposed relation, bound values, and supporting evidence \\
Task-end review & Candidate segments, entry snapshots, result evidence, and nearby operators & Local support judgment, retained steps, citations, and operator description \\
\bottomrule
\end{tabularx}
\caption{Inputs and outputs of TAM's model calls. Outcome extraction is
triggered for relevant unresolved outcomes; task-end review processes
recorded candidates.}
\label{tab:app_model_interfaces}
\end{table}

Task compilation occurs at initialization, acting follows the environment's
decision loop, and retrospective review occurs after termination. The cache
and the organization of available and missing inputs introduce no separate
model calls. Model-cost accounting includes compilation, acting, triggered
extraction, format repair, and retrospective review, together with the input
tokens of recalled images. The benchmark settings and context budgets are
reported in Appendix~\ref{app:protocol}; Appendix~\ref{app:computation}
analyzes the observed allocation of inference and actor context.

\subsection{Structured interfaces and representative prompts}
\label{app:concrete_interfaces}

The compiler returns a JSON object with a \texttt{goal\_slots} list.
Each requirement has a stable \texttt{schema} name and a
\texttt{goal\_kind}: \texttt{resource} denotes information to obtain,
and \texttt{done} denotes an observable GUI effect. The optional
\texttt{type} and \texttt{required\_fields} specify its value structure;
\texttt{requires} names prerequisite requirements.
\texttt{consumption\_role} distinguishes a target endpoint, rendered
content, decision input, lookup key, attachment, or verification target.
\texttt{entity\_pattern} and \texttt{retrieval\_cues} describe the object
and where to seek it. A conditional requirement names its
\texttt{branch\_predicate}; alternative decision requirements can also
name one another through \texttt{alternative\_to} and specify
\texttt{condition\_polarity}.
\texttt{effect\_constraints} preserves the task's required recipient,
content, attachments, and fields to be checked at completion.

For example, the compiler prompt contains the following complete small
output example. The date requirement provides content to the message
effect; a concrete task can add its recipient and other required inputs.
\begin{quote}\small
\begin{minipage}{\linewidth}
\begin{verbatim}
{"goal_slots": [
  {"schema": "arrival_date",
   "goal_kind": "resource",
   "type": "date"},
  {"schema": "sms_sent",
   "goal_kind": "done",
   "type": "done",
   "requires": ["arrival_date"],
   "effect_constraints": {
     "expected_render_schemas":
       ["arrival_date"]}}
]}
\end{verbatim}
\end{minipage}
\end{quote}

A stored binding records \texttt{schema}, \texttt{type},
\texttt{value\_norm}, and \texttt{state}; \texttt{entity\_key} identifies
its object, while \texttt{evidence\_ref} and \texttt{source\_step\_id}
retain the supporting artifact and source step.

Each update names an \texttt{op},
the displayed \texttt{schema}, a scalar \texttt{value} or structured
\texttt{fields}, an \texttt{evidence\_role}, and a short
\texttt{evidence\_text}. The roles are \texttt{source}, \texttt{derived},
\texttt{draft}, \texttt{coverage}, and \texttt{effect}.
A recalled-image update additionally cites the displayed
\texttt{reference\_evidence\_id}. These are proposals: the runtime
associates them with the task requirement and recorded observation.
An outcome-extraction call returns an \texttt{observations} list with
\texttt{schema}, \texttt{evidence\_ref}, \texttt{evidence}, and
\texttt{fields}. Its parser admits only listed requirement names,
the current post-action reference, and a nonempty visible witness.

The following excerpts are taken verbatim from the prompts used in our implementation;
the task-specific JSON packet and images follow the instructions.

\paragraph{Task compilation.}
\begin{quote}\small
``Infer the smallest set of goal slots required to determine task completion.''
``Use requires only for other goal-slot schemas that are semantic prerequisites
for this claim; do not include app names, actions, or concrete values.''
\end{quote}

\paragraph{Outcome extraction.}
\begin{quote}\small
``Report only the listed outcomes whose verification conditions are visibly
satisfied for the correct object and values. A draft, an attempt, or
disappearance of a form alone is not confirmation.''
``Use observations: [] when nothing is confirmed.''
\end{quote}

\paragraph{Retrospective review and abstraction.}
\begin{quote}\small
``Task success is not a local-evidence label. Keep supplied anchor/effect/step/evidence IDs.''
``Use inputs available at the retained start. Information acquired inside the
trace is internal work, not a prerequisite.''
``Compare that local input-to-outcome relation with nearby\_operators.''
``Use task roles, not instance values. The host retains each real route; do not average routes.''
\end{quote}
The reviewer returns one decision per candidate, containing its
\texttt{anchor\_id}, \texttt{effect\_schema},
\texttt{source\_validity} (supported, ambiguous, or contradicted),
\texttt{retained\_step\_ids}, \texttt{cited\_evidence\_ids}, and rationale.
For a supported result, an optional \texttt{operator\_update} contains
\texttt{same\_operator\_id} and short descriptions of its
\texttt{inputs}, \texttt{outcome}, and \texttt{method}.

\subsection{Candidate boundaries and retained interactions}
\label{app:concrete_boundaries}

An accepted local result supplies the candidate endpoint $e$.
For resource alignment, the implementation ranks recorded transitions up
to the detected endpoint by $H_i=4x_i+3n_i+2\max(0,z_i-y_i)$.
Here $x_i,y_i,z_i$ are the fractions of claim tokens appearing in the
action target, pre-action UI, and post-action UI, respectively;
$n_i$ is the fraction newly visible after the action. Claim tokens come
from the proposed values, fields, and evidence text, with generic
stop words removed. The highest-scoring transition is used, with later
steps breaking ties, only if $H_i\geq0.35$; backdating additionally
requires an exact claimed value in that transition's visible post-action
UI. Otherwise the detected endpoint is retained.
Let $r$ be the earliest recorded availability
step, before $e$, among the candidate's required inputs, and let $p$ be
the previous captured advancement step strictly before $e$. If a backdated endpoint is at or before the most recent advancement, that advancement does not constrain the start. The initial start is
$\max(r,p+1)$ when both exist, $r$ or $p+1$ when only one exists, and the
episode start otherwise. The candidate contains the bounded suffix of
this interval through $e$; the default setting retains at most eight
steps. An empty recorded slice falls back to the detected transition as a single-step candidate. Thus the reviewer receives a local candidate rather than choosing
an unrestricted interval from the full episode.

The reviewer chooses a nonempty subset of supplied step identifiers that
includes $e$, and cites evidence identifiers from the supplied packet.
The host validates both sets. It then closes the selection over observed
input support: input-entry steps not marked as making no progress, or selections of a result or
recipient bound to a required input, identify the earliest supporting
step. The retained route includes the nonempty, non-wait actions from
there through $e$. For acquisition without entry inputs, it also retains
the navigation/read prefix from the latest recorded app opening, or the
bounded prefix if no app opening is present. If required inputs have no
identified support step, the validated reviewer selection is retained.

Entry inputs are determined after this closure. The snapshot immediately
before the retained start supplies the available facts; subsequent
snapshots through the endpoint identify acquired or updated facts.
An input requirement is removed only if it was first acquired inside the
retained interaction and was absent at entry. A later retained start
therefore uses its own earlier-state snapshot. Missing history remains
unknown. When its application and activity are recorded, the GUI
realization's entry condition is taken from the first retained step's
observed pre-state; otherwise the original entry scope is retained.

\subsection{Operator identity, matching, and GUI readout}
\label{app:concrete_matching}

Retrieval compares operators learned from reviewed interactions with
requirements compiled from the current instruction. It checks the
required result and input roles before considering a recorded GUI route.
Our MobileWorld and SPA-Bench configurations share the reference
settings specified below. Task bindings and GUI context are recomputed
for each retrieval; execution evidence updates the reliability of each
operation and route.

The stored semantic record names its \texttt{operator\_kind} and
\texttt{operator\_family}, entry \texttt{requires}, expected
\texttt{effects}, and \texttt{verifies}. Its input profiles bind
requirement schemas to parameter roles; guard variants preserve optional
conditions. The \texttt{verification\_contract} names the result schema,
the required result fields, and the evidence relation. Separate GUI
records link to this semantic identifier and retain the entry context,
typed route, retained source steps, required/capability roles, and evidence
provenance.

\paragraph{Consolidation.}
The reviewer compares the retained operation's input-to-result relation
with the supplied nearby operators. It proposes an existing identifier
only when that relation is compatible; otherwise it returns null.
The host accepts a proposed merge only if the identifier occurs among
the supplied operators, the description is nonempty, and the result's
canonical family matches. Otherwise it stores the supported sample
separately. A merge updates the description and retains observed input
profiles, guards, and individual GUI realizations. The host does not
average their action sequences.

\paragraph{Schema canonicalization.}
Compiler names are lowercased and normalized with underscores.
Marker-based alias maps assign canonical result families and input roles. Effect families encode domain and
operation; resource families retain recognized role/domain and positional
qualifiers. Thus \texttt{SMS-Sent} and \texttt{text\_message\_sent} map
to \texttt{text\_send}, while \texttt{meeting\_time} and
\texttt{event\_time} map to \texttt{resource:event\_time}.
Input profiles additionally normalize declared roles, such as
\texttt{render} to \texttt{content}. Unrecognized effects use
\texttt{predicate:<schema>}; resources lacking a recognized role retain
\texttt{resource:<schema>}. In fallback schema comparison, qualifier
conflicts mean opposing marker groups (e.g., start/end,
source/destination, departure/arrival) or aggregate versus nonaggregate
markers; these receive zero similarity.

\paragraph{Semantic matching.}
Semantic matching tests whether an operator can supply the result of a
pending requirement. Result types and input roles constrain eligibility;
normalized schema comparison supports records with incomplete type metadata.
An explicitly typed source must match the target's resource-acquisition
or effect-realization kind and canonical result family. For records without an explicit family, TAM first derives a canonical
family from the schema name. Agreement with the target family gives a
match score of $1$. Otherwise, normalized schema comparison checks exact
identity, qualifier conflicts, effect-family or resource-role agreement,
and finally substring or token overlap.
The corresponding reference scores are $1$ for exact identity, $0$ for
qualifier conflicts, $0.94$ for a shared effect family, $0.86$ for a
shared resource role, $0.82$ for a substring match, and token Jaccard
otherwise. The reference fallback threshold is $\tau_{m,0}=0.80$.
Incompatible effect families and effect--resource pairs are capped at
$0.59$ and excluded from feedback-calibrated matching, so name overlap
cannot override a type conflict.
Compatible input profiles must agree on whether the operation consumes
inputs. When both profiles have role metadata, the source profile must
cover the target's required roles. The matched operation is rebound to the current requirement and its
inputs. Reuse preserves the input--result relation while substituting
the new task values; historical values remain provenance.
Profiles lacking role metadata retain the coarser check on whether inputs
are consumed.

\paragraph{GUI compatibility and reliability.}
GUI compatibility tests a recorded route against the live interface.
Each realization retains its own entry conditions, so a shared semantic
operation can support different interface-specific routes.
A route needs a supported semantic source, an eligible stored lifecycle,
a valid typed route identity, available target inputs, and the capability
roles required by those inputs. Typed actions must specify meaningful
roles: typing requires both target and value roles, opening an app a
value role, and clicking a target role. Known application or page-role
conflicts reject a route. An activity conflict also rejects it unless
the page role agrees. Otherwise applicability uses
application/activity agreement and token overlap with the
entry context; matching layout fingerprints provide additional support.

The GUI context records the application, activity, page role, and a
layout fingerprint. The fingerprint hashes the UI tree after masking
displayed text, content descriptions, and digits; without a UI tree, it
hashes the application, activity, and orientation. The first 16 hexadecimal
characters of SHA-256 identify the normalized layout.
After the conflict checks, the reference implementation scores the
available entry-context evidence:
\begin{equation*}
\begin{split}
a=\min\bigl(a_{\max},\;b+w_JJ+w_AA\\
\qquad\qquad+w_VV+w_hh\bigr).
\end{split}
\end{equation*}
Here $J$ is token Jaccard between live and stored entry descriptions;
$A,V\in\{0,1\}$ indicate application and activity compatibility;
and $h$ counts
matching known page-role and layout features. The reference settings use $b=0.08$,
$(w_J,w_A,w_V,w_h)=(0.42,0.28,0.18,0.08)$, and $a_{\max}=0.72$.
Application and activity compatibility test normalized name equality
or containment, followed by overlap or shared prefixes among informative
name tokens. Missing both an informative entry description
and an application pair gives $a=0$.

TAM maintains semantic and route reliability separately. Reuse trials are assessed during execution, and reviewed outcomes
update persistent counts at task end. An observed target advancement
supports the semantic operation, including when the actor adapts its
actions. Route success additionally requires following and completing
the stored route and reaching its destination. Reviewed counterexamples
update the corresponding semantic or route failure count. Ambiguous or
unattributed outcomes are censored; repeated observation identifiers
are counted once.
For count-backed records, both estimates use additive smoothing:
\begin{equation*}
r_j=\frac{s_j+\alpha_j}{s_j+f_j+\alpha_j+\beta_j},
\quad j\in\{\mathrm{sem},\mathrm{gui}\}.
\end{equation*}
Here $s_j,f_j$ are the corresponding positive and negative evidence
counts. The constants $\alpha_j,\beta_j$ are smoothing pseudocounts
that stabilize estimates from sparse observations. Their reference values
are $(\alpha_{\rm sem},\beta_{\rm sem})=(2,1)$ and
$(\alpha_{\rm gui},\beta_{\rm gui})=(1,1)$.

A semantic source must pass its reliability gate.
Recorded-route guidance additionally requires sufficient route
reliability and GUI applicability, together with the compatibility
checks above. The reference thresholds are
$\tau_{{\rm sem},0}=0.55$, $\tau_{{\rm gui},0}=0.45$, and $\tau_{a,0}=0.38$.
As observations accumulate,
changes in reliability can enable or suppress route guidance.
This adapts reuse to experience with each stored operation and route.

For feedback-calibrated retrieval, the four admission thresholds use
the reviewed local reuse outcomes accumulated within the same memory
sequence. Between tasks, each
threshold is set to the midpoint of the positive and negative mean
pre-execution scores, with its reference value contributing one fixed
pseudo-observation to each mean. Semantic outcomes calibrate the matching
and semantic gates; outcomes of attempted routes calibrate the route and
applicability gates. The reference value is used when either outcome
class is absent or the positive mean is not higher. Type, input-role,
and interface conflicts remain hard rejections.

If no route passes, a supported semantic operation can still provide
its inputs, result, and verification cue, with missing inputs marked
explicitly. This preserves task-level guidance when a recorded GUI
implementation is unsuitable.
Scoped negative evidence remains a rejection in either mode.

Eligible alternatives undergo two-axis $\epsilon$-Pareto selection,
followed by a bound that first favors new requirement coverage and then
distinct route families. For persistent graph retrieval, the two axes measure requirement
coverage and execution support:
\begin{equation*}
\begin{aligned}
C&=\lambda\sum_{q\in U_t}m_q\mathbf{1}[m_q\geq\tau_C],\\
E&=\eta a+(1-\eta)r_{\rm gui}.
\end{aligned}
\end{equation*}
Here $U_t$ contains the remaining requirements, and $m_q$ is the largest
schema-match score between $q$ and the candidate's target-bound results.
The reference settings use $\lambda=1.25$, $\tau_C=0.66$,
$\eta=0.62$, and $\epsilon=0.05$, retaining at most two alternatives. The result and verification sets
coincide on this path. Semantic-only advice has $E=0$; a persistent
candidate without typed requirement coverage is rejected.
A candidate is removed when another is at least as good minus
$\epsilon$ on both axes and better by more than $\epsilon$ on either.

\paragraph{Conflicting upstream evidence.}
The readout excludes unresolved or conflicted facts from available inputs,
so dependent requirements identify those inputs as missing. In the typed
state kernel, a pending effect whose inputs cease to be established becomes
unprepared; an attempted effect with a newly conflicted input becomes
ambiguous and calls for further evidence. A separately confirmed effect
retains its own observation-backed status when an upstream input becomes
conflicted.

\subsection{A complete illustrative operator for Figure~\ref{fig:tam_overview}}
\label{app:route_operator_example}

The following instantiates every component of the route-query operator
in Figure~\ref{fig:tam_overview}. Its role names and route description are
schematic, not an additional measured trajectory. Let $o$ and $d$ denote
the origin and destination and $\tau$ the observed driving duration.
\begin{description}
\item[Entry requirements $R_\kappa$.]
Origin and destination are available as the two route endpoints. The
driving duration is not an entry input.
\item[Expected advancement $\Delta_\kappa$.]
Acquire $\mathrm{driving\_duration}(o,d)=\tau$ for the specified route.
\item[Verification $V_\kappa$.]
A route-result observation shows driving mode, the bound origin and
destination, and the reported duration for that route.
\item[Object roles $B_\kappa$.]
The query's origin and destination and the verified route's endpoints
share the bindings $o,d$; the result binds $\tau$.
\item[Applicability $\Gamma_\kappa$.]
The requested result is a driving-route duration. This example has no
additional task branch guard; application and page conditions belong
to each realization.
\item[GUI realizations $\mathcal A_\kappa$.]
Each realization records its observed entry application/page, retained
interaction steps, and cited route-result evidence. A schematic route
opens the route query, enters $o,d$, selects driving mode, and inspects
the result. Different observed routes remain separate records.
\end{description}
The arrival-time calculation consumes the acquired duration and the
departure time; notifying Susan consumes that calculated time and the
recipient binding. These later results are outside the route-query
operator. On reuse, new endpoints replace $o,d$, and a new observation
must establish the new duration before it enters task state.

\section{Experimental Setup}
\label{app:protocol}

\subsection{Benchmarks and Observations}
\label{app:benchmarks}

\paragraph{MobileWorld.}
We evaluate the GUI-only partition of MobileWorld
\citep{kong2026mobileworld}. Based on the released task metadata
\citep{mobileworld2026release}, this partition contains 117 tasks
across 15 GUI applications, with 105 English and 12 Chinese
instructions. The application counts in the task metadata divide
these tasks into 55 single-app, 48 two-app, and 14 three-app workflows.
Tasks include communication, scheduling, document handling, file
management, and shopping. The GUI-only setting uses touch, text entry,
scrolling, navigation, application launch, waiting, and termination.
It excludes MCP tools and simulated-user clarification.

The observation interface supplies the current screenshot and the
available Android UI tree. Each method constructs its own history and
memory context from these observations. MobileWorld initializes the
device and application data for each task. Its native evaluators check
the requested result through task-specific answer checks, backend state,
device-local files, or application callbacks. Evaluation data are kept
separate from actor observations.

\paragraph{SPA-Bench.}
We use the English single-app Level~3 setting of SPA-Bench
\citep{chen2025spabench}. Based on the released task metadata
\citep{spabench2025code}, this split contains 50 tasks across
36 applications, with reference human trajectories comprising
9--18 annotated actions. These multi-stage single-app workflows complement
the cross-app demands of MobileWorld. The split is identified by the
English single-app task file and its Level~3 difficulty annotations;
task instructions are used as released. As specified in
Section~\ref{sec:experimental_setup}, TAM uses GPT-4o with Set-of-Mark
observations. Appendix~\ref{app:spabench_protocol} describes initialization,
action execution, final evaluation, and memory continuity for this setting.

\subsection{MobileWorld Implementation and Evaluation}
\label{app:benchmark}
\label{app:splits}

\paragraph{Interaction budget.}
All MobileWorld methods use Qwen3-VL-32B-Instruct and a budget of 50
controller rounds. A round is an externally counted action decision,
including a wait or context-management action. Some baseline decisions
expand into several primitive actions; these actions are recorded
separately while the decision contributes one controller round. Internal
memory operations that make no environment action do not consume a
round. Their model calls contribute to token expenditure.

All MobileWorld methods follow the same evaluation and stopping
protocol. The native evaluator checks the task after each executed GUI
action, and execution stops immediately when it confirms success. For
decisions containing multiple primitive actions, this check occurs after
each primitive action. Otherwise, execution ends when the agent
terminates or reaches the 50-round budget, and the final state is
evaluated. Any terminal answer is submitted before evaluation. The
evaluator's reference answers and internal state are not supplied to the
actor. SPA-Bench uses its own task-specific budget and post-execution evaluation
procedure, detailed in Appendix~\ref{app:spabench_protocol}.

\paragraph{Model and context settings.}
TAM uses temperature zero. Table~\ref{tab:app_policy_settings} records the
reference implementation's actor settings for both benchmarks. The actor receives the live
screenshot, task state, and recent action history. The Evidence Cache
can additionally supply the source images selected for a current
requirement, as described in Appendix~\ref{app:evidence_cache}.
Task compilation and operator review use the model interfaces summarized
in Appendix~\ref{app:implementation_limits}.

\begin{table}[t]
\centering\small
\setlength{\tabcolsep}{4pt}
\begin{tabularx}{\columnwidth}{@{}Yrr@{}}
\toprule
Setting & MobileWorld & SPA-Bench\\
\midrule
Temperature & 0 & 0\\
Actor output limit (tokens) & 8,192 & 1,024\\
Recent action records & 6 & 5\\
Operator context (tokens) & 900 & 900\\
Persistent operator capacity & 512 & 512\\
\bottomrule
\end{tabularx}
\caption{TAM reference implementation defaults. Baseline controllers
retain their own history management and role-specific output limits.}
\label{tab:app_policy_settings}
\end{table}

\paragraph{Persistent memory.}
The following memory lifecycle applies to both benchmarks. TAM builds its operator library from the task sequence without a separate
application-exploration stage. Task Advancement State and the Evidence
Cache are cleared at every task boundary. Reviewed operators persist
within the evaluation sequence and become available after the source
task ends. Thus a task uses the library available at its start; its own
retrospective review contributes to later tasks. The task manifest fixes
the sequence, and the library is maintained across execution batches.
Memory stores are separate for different methods and ablation
configurations. Methods with a preconstructed knowledge bank load that
bank before execution and retain their own retrieval mechanism.

\paragraph{Task outcomes and retries.}
Native evaluation determines task success: MobileWorld uses the per-action
and final checks described above, whereas SPA-Bench scores the recorded
trajectory after execution. A transport retry
resends the model request within the same decision and does not add a
GUI action. Interrupted attempts caused by unavailable services or an
invalid environment initialization are retained separately from completed
task outcomes. A task is retried only when its initial conditions and
memory boundary can be restored without duplicating side effects or
committed memory. The experiment ledger preserves the original attempt,
reported model usage, and reason for retry. Ordinary task failures are
not rerun to select a better outcome.

\subsection{SPA-Bench Implementation and Evaluation}
\label{app:spabench_protocol}

\paragraph{Environment and action interface.}
SPA-Bench tasks run in an Android emulator controlled through Android
Debug Bridge (ADB). Before each task, the adapter restores the device snapshot, performs the benchmark's task initialization, and
captures a fresh observation of the target application. Device and
capture-service preparation occurs before policy execution. Snapshot
restoration resets device-local state, external account state is a
separate task prerequisite. The adapter supports touch, long press,
dragging, text entry, scrolling, navigation, application launch, waiting,
and termination. These operations act on the live interface. The actor
receives the released instruction, current SoM observation, and its method's
history and memory, without access to reference trajectories or evaluator
key components.

\paragraph{Interaction budget and stopping.}
Following the released runner's default convention
\citep{spabench2025code}, task $i$ has a budget of $B_i=2g_i+1$
rounds, where $g_i$ is its annotated human action count. Our controller
reserves the final round for termination, allowing at most $2g_i$ GUI
interactions. Waiting and context-management decisions that issue an
environment action consume a round; internal memory computations and
request retries do not create additional GUI interactions. Execution
ends when the policy terminates or the budget is exhausted. An agent's
completion declaration closes the trajectory but does not establish
benchmark success.

\paragraph{Official evaluation.}
We use SPA-Bench's single-app evaluator with its \texttt{result\_only}
reasoning mode and \texttt{with\_action} trajectory representation
\citep{spabench2025code}. The evaluator first searches the recorded
screenshots for the task's textual key components using OCR. If this
screening stage passes, GPT-4o assesses the instruction and screenshot
trajectory with action annotations. The resulting official success
decision determines the binary task outcome. Evaluation is performed
after execution rather than after each action, and its annotations and
judgments are not fed back into online action selection.

\paragraph{Memory across tasks.}
The persistent-memory protocol in Appendix~\ref{app:benchmark} also
applies to SPA-Bench: TAM begins with an empty operator library and
retains reviewed operators in task order across execution batches.
Learning uses observed local results and retrospective review, without
using official evaluator annotations as memory content.
For the SPA-Bench ConAct baseline, we carry its
model-maintained explicit UI-memory dictionary across tasks as a declared
lifecycle extension to the released controller. Native memory add,
update, and delete operations are preserved, while folded action history
and step indices reset for each task. This extension adds no separate
summarization, retrieval, or learning call. Each method maintains its own
store over the declared sequence.

\paragraph{Comparison protocol.}
Our TAM and ConAct evaluations use the same GPT-4o backbone, task
instructions, observation interface, initialization procedure, and
task-specific interaction budgets. Each method retains its own memory
representation and context management. Table~\ref{tab:spabench} also
includes the first seven baselines and GUI-explorer as published
reference results from \citet{xie2025guiexplorer}. Their reported inputs
and backbones are retained; the shared execution protocol applies to our
implementations rather than retroactively to those published runs.
The MobileWorld baseline adaptations in
Appendix~\ref{app:baseline_adaptation} concern
Table~\ref{tab:mobileworld}.

\paragraph{Evaluation sequences and aggregation.}
\label{app:spabench_aggregation_design}
The design uses all 50 official English single-app Level~3 tasks in two
complete evaluation sequences.
Each task receives one scored attempt per
sequence.
TAM and ConAct share the task orders, generated from the
lexicographically sorted task identifiers with order seeds 0 and 1.
These seeds determine task order, they do not control model sampling.
Each method starts each sequence with an empty persistent store.
Task-local state and the environment reset between tasks, while persistent memory
accumulates within the sequence.
For method $m$, let $y_{mri}\in\{0,1\}$ be the official outcome for task
$i$ in sequence $r$.
The reporting rule is
\begin{equation}
\begin{aligned}
S_{mr} &= \frac{100}{50}\sum_{i=1}^{50}y_{mri},\\
\bar S_m &= \frac{1}{2}\sum_{r=1}^{2}S_{mr},\\
s_m &= \sqrt{\frac{1}{2}\sum_{r=1}^{2}(S_{mr}-\bar S_m)^2}.
\end{aligned}
\label{eq:spa_reporting_design}
\end{equation}
Each of TAM and ConAct contributes 100 scored task episodes
(50 tasks in each of two complete sequences), giving 200 scored
episodes across the two methods.
All tasks have equal
weight, and percentages are rounded only after aggregation.
Variation across the two sequences describes the combined effects of task order
and execution variability.

\paragraph{Memory updates and interrupted attempts.}
Under this design, task-end learning runs after every valid episode,
including failures, using the review and storage rules in
Section~\ref{sec:operator_learning}.
The terminal binary verdict can participate in the disclosed task-end
review, hidden reference answers and evaluator annotations are not stored
as task knowledge.
Operators become available after the source task ends.
Task failures, premature termination, and exhausted action budgets count
as zero and retain their place in the sequence.
Only external
infrastructure faults that prevent valid execution or scoring permit a
replacement attempt.
If that state cannot be restored, the sequence restarts from empty memory
in the same order.
Persistent infrastructure failure is reported as an incomplete
sequence rather than a reduced task denominator.

\subsection{Component Ablations}
\label{app:ablation_protocol}

The component study in Table~\ref{tab:ablation} varies the enabled memory
components while retaining the base policy, decoding settings,
observation adapter, action interface, recent-history window, and
interaction budget. All configurations with Task Advancement State use
the same task compiler and state-update procedure.
Table~\ref{tab:app_component_switches} specifies the interventions.

Disabling the Evidence Cache removes recalled source images while
retaining established bindings and ordinary history. Disabling the
Advancement Operator Graph removes persistent retrieval and task-end
operator learning while retaining online state updates. Each
configuration maintains a separate library over the same task sequence.
The flat-state comparison in Appendix~\ref{app:state_control} is a
separate intervention on the organization of task-local information.

\begin{table}[t]
\centering\small
\setlength{\tabcolsep}{3pt}
\begin{tabular}{@{}lccc@{}}
\toprule
Configuration & State & Cache & Graph\\
\midrule
Base agent & -- & -- & --\\
Advancement State & $\checkmark$ & -- & --\\
State + Evidence Cache & $\checkmark$ & $\checkmark$ & --\\
State + Operator Graph & $\checkmark$ & -- & $\checkmark$\\
Full TAM & $\checkmark$ & $\checkmark$ & $\checkmark$\\
\bottomrule
\end{tabular}
\caption{Component ablations. State includes task compilation and
advancement updates. Graph includes both retrieval and task-end learning.}
\label{tab:app_component_switches}
\end{table}

\subsection{Metric Definitions}
\label{app:metrics}

Let $N$ be the number of evaluated task instances, $s_i\in\{0,1\}$ the
native-evaluator outcome, and $a_i$ the number of controller rounds up to
the stopping point. Under the common MobileWorld protocol, a successful
trajectory ends at the first evaluator-confirmed success; an unsuccessful
trajectory ends at agent termination or the round budget. We report
\begin{equation}
\begin{aligned}
\mathrm{Success} &= \frac{100}{N}\sum_{i=1}^{N}s_i,\\
\mathrm{Steps} &= \frac{1}{N}\sum_{i=1}^{N}a_i.
\end{aligned}
\end{equation}
Steps includes both successful and unsuccessful tasks and is read
alongside Success.

\paragraph{Efficiency--Quality Area.}
Following the fixed-budget formulation of MMBench-GUI
\citep{wang2026mmbenchgui}, we compute
\begin{equation}
\mathrm{EQA@}B
= \frac{100}{N}\sum_{i=1}^{N}
s_i\max\!\left(0,1-\frac{a_i}{B}\right),
\label{eq:app_eqa}
\end{equation}
with $B=50$ for MobileWorld. EQA@50 uses the same stopping point as Steps.
Each successful trajectory contributes according to its completion
length; a failed trajectory contributes zero. EQA@50 therefore rewards
success achieved with fewer decisions.

\paragraph{Lifecycle Tokens per Solved Task.}
Let $T_{\mathrm{build}}$ be the model-token expenditure for constructing
memory before evaluation, and $T_{\mathrm{online}}$ the expenditure during
the task sequence, including actor, state-update, retrieval, reflection,
review, and repair calls. Then
\begin{equation}
\mathrm{LTS}=
\frac{T_{\mathrm{build}}+T_{\mathrm{online}}}{\sum_i s_i}.
\label{eq:app_lts}
\end{equation}
The numerator includes input and output tokens from successful and
failed attempts. Execution calls are counted through the stopping point.
Subsequent task-end reflection, review, and memory updates are also
included. Vision and cached-input tokens follow the provider's
reported usage. Construction shared across tasks is counted once in the
numerator and amortized over the solved tasks. We report LTS in millions
of tokens. As indicated in Table~\ref{tab:mobileworld}, \mbox{StructAgent} and
GUI-explorer have lower-bound LTS values because the available ledgers
exclude their original knowledge-bank construction.

The mechanism analyses use the task, field, operation, or episode units
defined in the Analysis Protocols and Sample Definitions in
Appendix~\ref{app:aggregate}. Those diagnostic
populations are separate from the benchmark-level denominators above.

\section{Baseline Adaptations}
\label{app:baseline_adaptation}

All MobileWorld methods use the evaluation and stopping protocol in
Appendix~\ref{app:benchmark}. The integrations retain each method's memory
representation, update procedure, and internal model roles. Adapters
translate their actions to the common mobile interface and expose the
current screenshot, available UI-tree information, and application-launch
operation. Internal calls contribute to the lifecycle accounting in
Appendix~\ref{app:metrics}.

\paragraph{AWM \citep{wang2025awm}.}
AWM induces workflow descriptions from successful trajectories and
replaces task-specific arguments with variables. The online adapter
updates the workflow library at task boundaries and includes its text
in the actor prompt in source order, up to 12,000 characters. Induction
requires at least two actionable steps and has a 2,048-token output
limit. The actor retains the task's previous action history and uses a
1,024-token output limit. Workflow descriptions preserve application
and affordance names while abstracting instance values and omitting absolute coordinates.

\paragraph{Mobile-Agent-E \citep{wang2025mobileagente}.}
The integration retains the Manager, Operator, Action Reflector,
Notetaker, and the experience reflectors for tips and shortcuts. Its
perception pipeline combines OCR and icon descriptions with available
UI-tree controls. After an action, the agent observes the result,
reflects on the action, and updates its notes. In the evolution setting,
tips and shortcuts persist across tasks, while working notes reset.
The optional experience selector is disabled, so available tips and
shortcuts are supplied directly. Role responses have a 2,048-token
limit. Shortcut expansion is preserved, with its atomic actions logged
separately from the controller decision.

\paragraph{\mbox{StructAgent} \citep{wu2026structagent}.}
The adapter retains structured task state, planning, perception,
milestone verification, completion auditing, and recovery. It uses the
released fixed experience bank with up to ten retrieved items and at
most five planner images. The action adapter translates supported
operations to Android interactions; desktop shell, browser-control,
and office tools are unavailable in the GUI-only environment. The
default role output limit is 1,500 tokens, with role-specific requests
capped at 8,192. The original experience bank remains fixed during the
task sequence.

\paragraph{GUI-explorer \citep{xie2025guiexplorer}.}
The adapter preserves exploration-derived knowledge, multimodal
retrieval, model-based ranking, action selection, and step
summarization. The live screen queries the knowledge base through its
embedding and FAISS pipeline. Retrieval requests one candidate per
image query at a similarity threshold of 0.9. The ranker selects
knowledge relevant to the current instruction, and step summaries
retain execution history. Action selection and summarization use a
1,000-token response limit. The Device interface maps selected actions
to MobileWorld while retaining this retrieval and execution loop.

\paragraph{MemGUI-Agent (ConAct) \citep{liu2026memgui}.}
We adapt the MemGUI-Agent implementation released with MobileWorld.
It retains the five-part tagged response format, structured context,
history folding, and explicit context-management operations. The
actor receives one current screenshot with the shared UI-tree and
application-launch extensions; its own context operations manage
earlier information. A context-management operation executes a wait
and consumes one controller round. Context resets at each task
boundary. The response limit is 8,192 tokens to accommodate the tagged
sections and rewritten context. Parsing permits up to three attempts
on unchanged messages; these attempts are distinct from transport
retries and do not introduce extra environment actions.

\section{Additional Results and Analysis}
\label{app:aggregate}

The following analyses examine information retention, input preparation,
operator learning and reuse, and resource allocation. They draw on
benchmark metadata, historical execution archives, paired state controls,
and readout tests at fixed checkpoints. The benchmark
comparisons measure end-to-end completion, these diagnostics examine the
specific behaviors that task advancement is designed to support.

\subsection{Benchmark Coverage of the Diagnostics}
\label{app:workflow_demands}

\begin{figure*}[t]
\centering
\includegraphics[width=\textwidth]{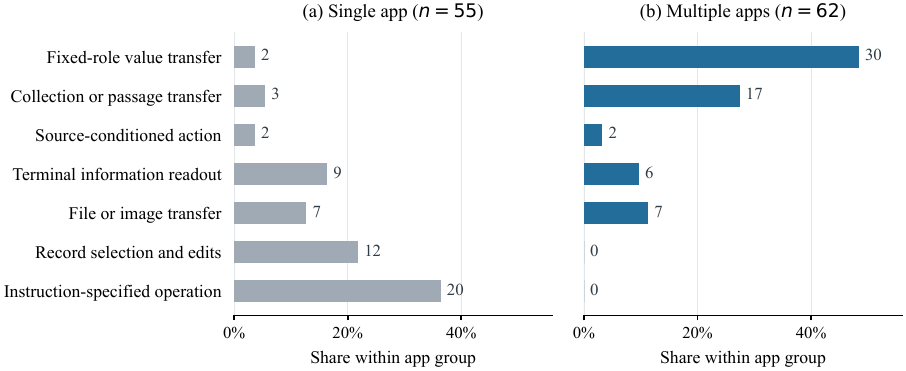}
\caption{Information requirements in the MobileWorld GUI-only inventory.
Bars show within-group percentages and labels give task counts. Categories
describe each task's principal requirement. Application counts follow
benchmark metadata.}
\label{fig:information_demands}
\end{figure*}

Field-level diagnostics require both an acquired value and an identifiable
destination for its later use. We classify all 117 MobileWorld GUI-only
tasks by the principal information requirement in their instructions and
official definitions (Figure~\ref{fig:information_demands}). The classification
identifies which tasks support this diagnostic.

The census identifies 52 tasks whose principal requirement is to transfer
acquired text or numerical content to a later GUI operation, comprising
47 multi-app tasks and five single-app tasks. Tasks involving final answers,
decisions based on source information, file transfers, or navigation require
different evaluation criteria. A source-to-target field test therefore
applies to a particular information requirement, rather than to every
multi-app task.

These 52 tasks define the benchmark scope of source-to-target field
scoring. The main-text audit uses a reference catalogue covering 21 of
them, comprising 14 fixed-role and seven collection tasks. Its 52 shared
values provide identical field references across the five methods.
The catalogue covers a subset of the applicable tasks.

\subsection{Preserving Information beyond Its Source}
\label{app:value_persistence}

We measure retention after information has entered state. This requires
an identifiable acquisition followed by an opportunity to retain the value.
We therefore track values that match both their cited read-only UI source
and the state supplied to the actor. Instruction constants and echoes of
the agent's own input are excluded. This defines a diagnostic of acquired
information with a verifiable textual source.

\begin{figure}[t]
\centering
\includegraphics[width=\columnwidth]{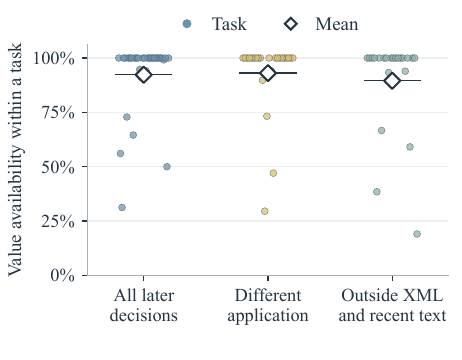}
\caption{Availability after acquisition. Tasks contribute when they have
qualifying later decisions: 32 overall, 24 in a different application
from the source, and 22 outside current XML and recent text. Points show task rates and
diamonds their means. Horizontal jitter separates overlapping points.}
\label{fig:value_persistence}
\end{figure}

Scanning all 756 valid archived TAM-family episodes yields 202 bindings
across 32 tasks, where a binding associates a task role with a value.
These 32 tasks contribute every acquisition satisfying the above rule.
The 198 bindings with subsequent decisions yield 3,751 value--decision
pairs. Every later decision is scored, including those where the value
has disappeared, regardless of episode success. Availability is the
fraction of these pairs in which the value remains in state, averaged
equally across tasks. It reaches 92.3\% overall, 93.1\% in an application
different from the source, and 89.6\% when the value is absent from both current XML and
recent text (Figure~\ref{fig:value_persistence}).

\begin{figure}[t]
\centering
\includegraphics[width=\columnwidth]{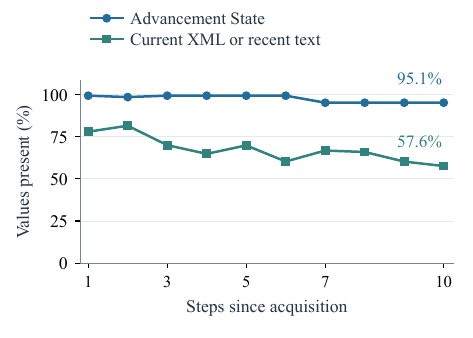}
\caption{Ten-decision follow-up of every binding with a complete observation
window, yielding 106 bindings from 24 tasks. Both curves use these same
values throughout, with equal task weights, and compare text locations
within TAM executions.}
\label{fig:fixed_followup}
\end{figure}

For the temporal comparison, we require ten subsequent recorded decisions
per binding so that every point follows the same values. Otherwise,
episodes ending earlier would change the composition of the curve.
All 106 bindings meeting this duration requirement enter the analysis,
covering 24 tasks. Values remain in the denominator even when they
disappear. After ten steps, 95.1\% remain in state, compared with 57.6\%
present in current XML or recent text (Figure~\ref{fig:fixed_followup}).
These results show that TAM preserves acquired information across
application changes and extended action sequences. Advancement State
keeps earlier results available to later decisions even when they are
absent from the current XML and recent text.

Scores measure literal retention in primary state, excluding quoted
evidence. Observation ends with the episode. A value may disappear after
replacement or completed work. The shared protocol specifies matching
rules and longer follow-up windows.

\subsection{Preparing Inputs for an Operation}
\label{app:state_control}

We test whether state organization brings an operation's source-derived
inputs together before entry. The ten-task panel was chosen by task
structure to cover cross-application dependencies, one-to-many coordination,
document grounding, and terminal effects. Flat state retains facts and
unfinished requirements while removing prerequisite links and associated
action guidance. Both conditions disable persistent retrieval and learning.
We examine all catalogued destination operations in this panel. Nine tasks
contribute 23 fields grouped into ten operations. The remaining task requires
a decision without field entry. Two saved revisions provide 20 operation
instances.

\begin{figure*}[t]
\centering
\includegraphics[width=\textwidth]{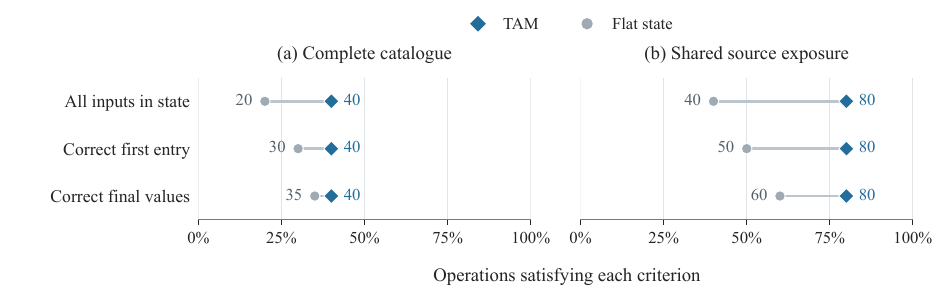}
\caption{Operation preparation and use in (a) the full diagnostic panel and
(b) the shared-source subset of ten instances. The 20 full-panel instances comprise
ten operations evaluated in two saved revisions. Each line joins TAM
and flat-state percentages on the same set of operations. A criterion
requires every audited parameter to satisfy it.}
\label{fig:input_preparation}
\end{figure*}

An operation is prepared when all its audited, environment-derived target
values appear together in state before the first entry attempt. TAM prepares
40\% of operations versus 20\% for flat state. Correct first entry requires
all audited fields to be correct on their initial entry. Final accuracy
allows repairs, reaching 40\% and 35\%, respectively
(Figure~\ref{fig:input_preparation}(a)).

The shared-source comparison examines input organization in workflows
where both agents encounter the required source information. All operations
meeting this condition are included, yielding ten instances from five tasks
across the two revisions, including invoice cases that both agents fail.
TAM prepares
80\% versus 40\% for flat state and achieves correct first entry for 80\%
versus 50\% (Figure~\ref{fig:input_preparation}(b)).

These ten instances contain the 18 field instances in the main-text paired
analysis. Using the first entry after complete source exposure as the
checkpoint gives the same results. Across the six multi-input instances,
TAM prepares 66.7\% versus 33.3\% for flat state, while final correctness
is 66.7\% for both. The paired comparisons show the value of organizing
facts around the operations that need them. TAM more often assembles
inputs before entry and achieves more correct first attempts when both
agents have encountered the source information.

\subsection{Access to Captured Cross-Application Information}
\label{app:readout}

We test access to information already recorded in TAM's state after an
application change. The same questions compare reading this information
from memory with recovering it from the interaction history.

A readout question requires a source-linked fact in state at a checkpoint
in a different application from its source. This tests access to captured
information after a context change. The archive yields 26 distinct eligible claims. A fixed-seed
selector retains at most two per task to limit repeated contributions
from one workflow, yielding 23 claims. We score all 20 with resolvable
source references, spanning 13 tasks and 14 checkpoints, including an
incorrect stored value. The claim selector does not inspect episode
success or probe answers. Three claims have unresolved object, quantity,
or temporal references. The shared protocol details the selection.

In each readout test, the same model answers the same question from the
same current observation under five context conditions: TAM state,
ordinary notes, full XML/action history, history with sampled images,
and recent context. Notes are generated from the
interaction prefix before the query is supplied, without access to TAM
state or the reference answer. A query is correct only when every requested
field matches its source-checked reference after format normalization.
We average queries within each task before averaging across tasks, and
also report the query mean.

\begin{table}[!htbp]
\centering\small
\setlength{\tabcolsep}{3pt}
\begin{tabular}{@{}lrrr@{}}
\toprule
Context & Task mean & Query mean & Tokens \\
\midrule
TAM state & \textbf{96.2} & \textbf{95.0} & 4,040 \\
Ordinary notes & 80.8 & 80.0 & 4,003 \\
Full XML/actions & 92.3 & 90.0 & 6,439 \\
XML + sampled images & 88.5 & 85.0 & 19,347 \\
Recent context & 26.9 & 25.0 & 3,883 \\
\bottomrule
\end{tabular}
\caption{Readout accuracy (\%) on the same 20 queries and mean input
tokens per readout. Task means give equal weight to each of the 13 tasks.
Query means give equal weight to each query.}
\label{tab:readout_context}
\end{table}

\begin{figure}[t]
\centering
\includegraphics[width=\columnwidth]{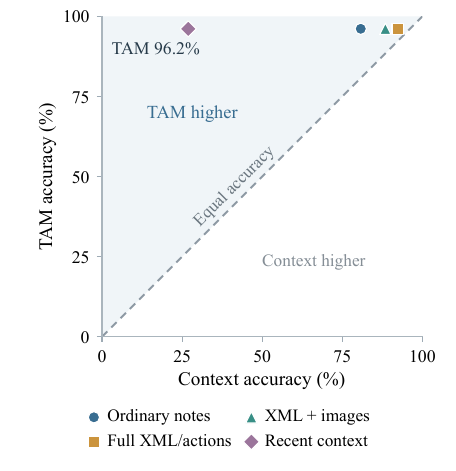}
\caption{Readout accuracy by context. Each point compares an
alternative context with the same TAM state condition. Axes give
task-averaged accuracy on identical questions. Points above the dashed
diagonal favor TAM.}
\label{fig:readout_pairs}
\end{figure}

TAM achieves the highest observed accuracy under both weightings
(Table~\ref{tab:readout_context}), with input size close to ordinary notes
and 37.3\% fewer input tokens than full textual history. Under the same
task weighting, TAM state reduces readout error by 80.0\% relative to
notes, 50.0\% relative to full textual history, and 66.7\% relative to
visual history. These are relative error reductions.
TAM's accuracy advantage is largest over recent context and ordinary
notes (Figure~\ref{fig:readout_pairs}).

\begin{table}[t]
\centering\small
\setlength{\tabcolsep}{3pt}
\begin{tabular}{@{}lrrr@{}}
\toprule
Context & Incomplete & Time & Amount \\
\midrule
TAM state & 0 & 0 & 1 \\
Ordinary notes & 3 & 1 & 1 \\
Full XML/actions & 0 & 1 & 1 \\
XML + sampled images & 0 & 2 & 1 \\
Recent context & 22 & 0 & 1 \\
\bottomrule
\end{tabular}
\caption{Errors among the same 28 requested fields per context.
Incomplete fields contain missing or partial answers. Time and amount
errors concern derived values. All observed field errors are included.}
\label{tab:readout_error_types}
\end{table}

The field-level breakdown distinguishes incomplete answers from incorrect
derived values (Table~\ref{tab:readout_error_types}). TAM, full textual
history, and visual history achieve 100\% field accuracy on multi-field
queries, compared with 76.9\% for notes. The additional errors
under full and visual history concern derived times. All five contexts
repeat one incorrect amount already present in the interaction prefix.

TAM combines accurate access to captured information with compact
context. Its state supports more reliable cross-application readout
than ordinary notes at a similar input size, while using substantially
fewer input tokens than full textual history.

\subsection{Quality of Local Learning Signals}
\label{app:local_learning_quality}

\begin{figure*}[t]
\centering
\includegraphics[width=\textwidth]{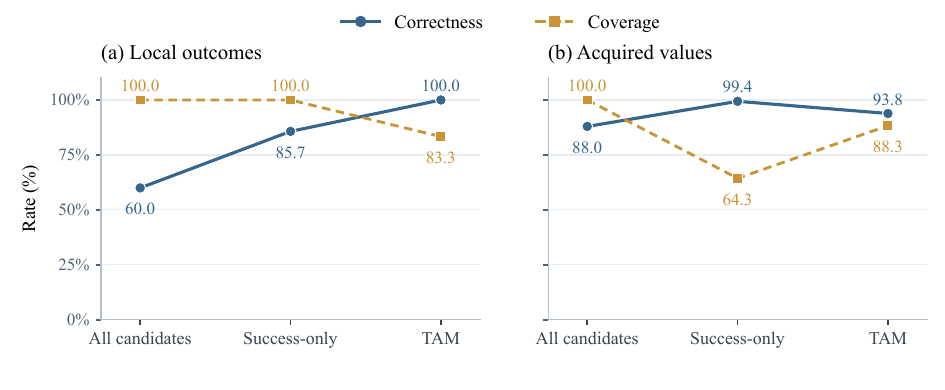}
\caption{Correctness and coverage of retained learning material.
Correctness is the supported fraction of retained items. Coverage is the
retained fraction of all supported items. Policies share the same pool
within each panel. Source-value weights give tasks equal representation
before selection.}
\label{fig:local_selection_quality}
\end{figure*}

\begin{table}[t]
\centering\small
\setlength{\tabcolsep}{3pt}
\begin{tabular}{@{}llrr@{}}
\toprule
Weighting & Selection & Correctness & Coverage \\
\midrule
Task & All & 88.0 & 100.0 \\
 & Success-only & 99.4 & 64.3 \\
 & TAM storage & 93.8 & 88.3 \\
\midrule
Reference & All & 84.2 & 100.0 \\
 & Success-only & 99.0 & 63.6 \\
 & TAM storage & 92.3 & 81.6 \\
\bottomrule
\end{tabular}
\caption{Correctness and coverage (\%) under fixed weights on the same
234 complete fields. Weights
are the inverse of each group's size in the full candidate pool. Supported
values remain in the coverage denominator when their group retains none.}
\label{tab:source_value_weighting}
\end{table}

We apply three policies to the same recorded candidate pools, retaining
all candidates, retaining only candidates from successful episodes, or
following TAM's storage decisions. A candidate pairs a claimed local result
with its interaction evidence. Tasks contribute through the learning
candidates they produce. Each test includes all candidates meeting its
reference conditions below. Correctness measures the quality of retained
material, while coverage measures how much correct material is retained.

\paragraph{Local outcomes.}
To assess local outcome judgments independently of terminal success, we
include all 20 admitted proposals made without terminal evaluator feedback.
Proposals nominated by the evaluator already use the terminal signal being
compared and are accounted for separately.
TAM's retained outcomes are 100\% supported and cover 83.3\% of the
supported pool (Figure~\ref{fig:local_selection_quality}(a)). Success-only
selection gives 85.7\% correctness. It admits premature completion claims
from a workflow that succeeds later. TAM excludes these contradicted
claims. Missing supported outcomes result from review-output processing
failures, which remain included in the coverage calculation.
Local verification thus improves the reliability of outcome evidence
selected for operator learning.

\paragraph{Acquired values.}
Value-level scoring compares a specified resource field with an external
reference. Within the source-reference catalogue, we include every
admitted field with a complete, unambiguous value claim and a determinate reference
in benchmark fixtures, injected messages, or source documents. This yields 234 fields across 20 tasks, including correct
and incorrect values. Incomplete claims and roles without resolved references
remain in the full candidate accounting
(Table~\ref{tab:full_source_accounting}).

To prevent candidate-rich tasks from dominating the comparison, we fix
weights before selection so that each task contributes equally to the
common pool. TAM retains 88.3\% of correct material, compared with 64.3\%
for success-only selection. Correctness among retained values is 93.8\%
and 99.4\%, respectively (Figure~\ref{fig:local_selection_quality}(b)).
TAM combines higher retained-value correctness than keeping all candidates
with broader coverage of correct material than success-only selection.
This expands the evidence available for operator learning, including
supported progress from unsuccessful workflows.

\paragraph{Sensitivity to repeated values.}
Table~\ref{tab:source_value_weighting} also gives equal weight to reference
groups defined by task, resource role, and reference value. Correctness
and coverage are ratios of weighted counts. Supported values remain in
the coverage denominator even when their group retains none. TAM's
coverage advantage persists under both weightings, showing that the gain
holds when either tasks or repeated source values receive equal weight.
The shared protocol also reports the unweighted comparison.

\subsection{Learning from Incomplete Workflows}
\label{app:failed_workflows}

We examine when correct local information first appears in workflows
that ultimately fail. This timing question requires at least one
source-checked correct record. We include every failed execution meeting
this condition in the source-value audit, yielding 29 executions across
16 tasks. Records contribute whether or not they are retained for learning.
Positions are normalized by the number of executed decisions, and tasks
receive equal weight.

\begin{figure}[t]
\centering
\includegraphics[width=\columnwidth]{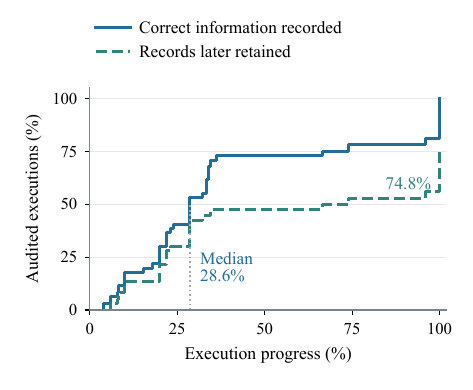}
\caption{Timing within failed workflows containing a source-checked
correct candidate, with tasks weighted equally. Both curves use
the original recording time. The dashed curve tracks records subsequently
retained for learning, with the same execution denominator.}
\label{fig:partial_progress_failure}
\end{figure}

The median first-record position is 28.6\% of execution, and 72.9\% of
audited executions contain a correct record by their midpoint
(Figure~\ref{fig:partial_progress_failure}). Correct material from 74.8\%
of these executions is eventually retained for learning. The dashed
curve locates those retained records within their source workflows.
The timing results show that useful learning material can arise early
in a workflow that ultimately fails. TAM retains this partial progress
for operator learning, extending the value of an execution beyond its
terminal outcome.
Table~\ref{tab:partial_progress_types} breaks down retention and timing
by information type.

\begin{table*}[t]
\centering\small
\setlength{\tabcolsep}{5pt}
\begin{tabular*}{\textwidth}{@{\extracolsep{\fill}}lrrrr@{}}
\toprule
Information type & \shortstack{Correct material\\retained}
& \shortstack{Executions yielding\\retained material}
& \shortstack{Correct record\\by midpoint}
& \shortstack{Median first-record\\position} \\
\midrule
Amounts and quantities & 100.0 & 100.0 & 50.0 & 34.6 \\
Contact and address details & 66.7 & 66.7 & 62.5 & 33.3 \\
Dates and times & 79.4 & 77.9 & 87.5 & 20.0 \\
\midrule
All (deduplicated) & 74.7 & 74.8 & 72.9 & 28.6 \\
\bottomrule
\end{tabular*}
\caption{Retention and timing in failed workflows containing correct
candidates (\%). Rates give equal weight to tasks within each information type. Positions are fractions of executed decisions, and the median uses the
task-weighted timing distribution. Types may share executions,
and the overall row uses their deduplicated union.}
\label{tab:partial_progress_types}
\end{table*}

Operators linked to 15 retained candidates subsequently appear in other
tasks' actor prompts. Their recorded lineage connects these deliveries to
ten correct destination uses. The shared protocol accounts for all delivered
cases. These traces document delivery and subsequent behavior, while the
main-paper ablation evaluates persistent operators as a whole.

\begin{figure*}[t]
\centering
\includegraphics[width=\textwidth]{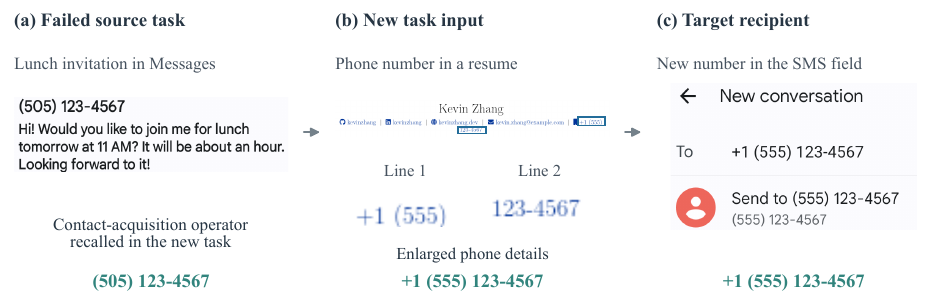}
\caption{A recorded source-to-target chain. The failed source workflow
correctly obtains a phone number. Its contact-acquisition operator is
later recalled without the old number or GUI route. The target actor reads
a different number from a resume and enters it as the SMS recipient.
Panels use cropped recorded screenshots. Enlarged details in (b) show
the phone number wrapped across two lines.}
\label{fig:failed_source_case}
\end{figure*}

Figure~\ref{fig:failed_source_case} illustrates one such chain. The source
correctly reads a sender's number from a lunch invitation but fails to
create the calendar event. Its operator is recalled for an interview
invitation, where the actor reads a different number from a resume, enters
it as the recipient, and completes the task.
This chain demonstrates how supported progress from an incomplete
workflow enters memory and becomes guidance for a different task.

\subsection{Using Recalled Experience with New Inputs}
\label{app:new_inputs}

We examine how recalled acquisition experience is used in a later task.
A reuse opportunity begins when an operator reaches the actor. Scanning
all 756 archived executions yields 206 such cases across 60 tasks, counting
each operator once per execution. Of these, 165 concern information
acquisition. We follow the first subsequent binding for the recalled
resource, including incorrect and unresolved values. All 66 deliveries
with such a binding enter the value audit, yielding 86 fields across
21 tasks. The other 99 deliveries remain in the coverage account. Task references support 65 of the
86 fields, contradict two, and leave 19 unresolved.

\begin{figure*}[t]
\centering
\includegraphics[width=\textwidth]{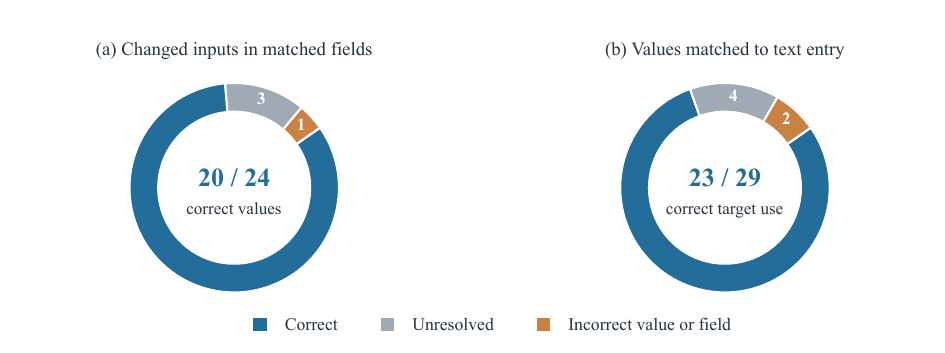}
\caption{Two diagnostic subsets of the full 86-field target audit.
(a) The 24 newly acquired inputs with a changed reference and comparable
source and target fields. (b) The 29 fields matched to executed text entry.
The two orange cases in (b) comprise one incorrect value and one correct
value entered only in an unrelated field. Unresolved cases remain visible.}
\label{fig:new_input_and_target_use}
\end{figure*}

\paragraph{Adapting to changed values.}
We test inputs that the target task requires the agent to obtain from
the environment. Source and target must refer to the same information
requirement and semantic field, with comparable value types. For example,
a phone number is compared with a phone number, and an event's start time
with another start time. A changed reference distinguishes adaptation to the target task from
repeating a source value. We include every field whose target reference
differs from all comparable source values. Of the 24 qualifying fields,
20 match the new reference, one is incorrect, and three remain unresolved
(Figure~\ref{fig:new_input_and_target_use}(a)).

\paragraph{From a binding to a destination.}
Literal text entry links a recorded value to a destination control.
We inspect every bound value matched to an executed text-entry action
and check its destination in subsequent XML or screenshots. Of 29 matched
fields, 23 (79.3\%) reach the intended destination with the correct value
(Figure~\ref{fig:new_input_and_target_use}(b)). One enters an incorrect
value, one occurs only in an unrelated message-search field, and four
lack a resolved reference. The correct uses span 22 actions in 22 episodes
across seven tasks. Of the 23 correct field uses, 21 follow semantic
guidance without an executable GUI route. The other 57 fields have no matching text-entry action and are
outside this destination test, which does not cover other input modes
such as time pickers. The two tests examine overlapping subsets of the same 86 fields.
Together, the tests show that TAM combines reusable acquisition guidance
with information obtained for the current task. The transfer traces
demonstrate correct binding and destination use with new task values,
supporting the reuse of experience across changing inputs.

\subsection{Inference and Context Allocation}
\label{app:computation}

We examine how recorded inference is divided between the actor and
additional model calls, and how much actor text is devoted to state and
operator guidance. A shared actor prompt format gives the state-and-guidance
block consistent boundaries, making its text share comparable across
executions. We include every valid archived execution using this format,
yielding 250 episodes across 96 tasks. Repeated executions are averaged
within each task. The records include five archived memory settings,
detailed in the shared protocol.

\begin{figure}[t]
\centering
\includegraphics[width=\columnwidth]{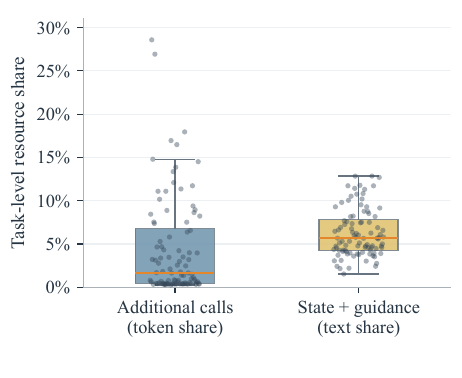}
\caption{Resource allocation for all 96 tasks with valid archived
executions using the shared actor prompt format. Shares refer to tokens
used by additional calls and actor text devoted to state and operator
guidance. All task values are shown. Boxes span the interquartile range,
orange lines mark medians, and whiskers extend to observations within
1.5 interquartile ranges of the box.}
\label{fig:context_compute}
\end{figure}

For each task, we average the fraction of recorded tokens used by
additional model calls across its executions. Across tasks, the median
is 1.7\% (interquartile range 0.4--6.8\%). For actor context, we first
take the median fraction of prompt characters devoted to state and
operator guidance within each execution, then average these episode
medians within each task. The median across tasks is 5.7\% (4.3--7.8\%), as
shown in Figure~\ref{fig:context_compute}. This measure covers 7,107
saved actor prompts.

Across configurations, the median token share of additional model calls
ranges from 0.5\% to 13.4\%, and the median text share from 4.1\%
to 6.3\%.

These profiles show that TAM provides explicit task state and reusable
guidance through a compact addition to actor context. The small median
share of tokens used by additional calls complements this representation,
supporting the practicality of integrating memory into ongoing execution.

\subsection*{Analysis Protocols and Sample Definitions}
\label{app:analysis_protocols}

\paragraph{Sources and links to the main results.}
The source labels in Table~\ref{tab:analysis_cohorts} distinguish benchmark
metadata (B), historical executions (H), paired state controls (P), and
fixed-checkpoint readout tests (Q).
Historical executions cover multiple
prompt and memory configurations.
They document observed behavior under
the recorded configuration of each episode.
The paired state study holds
its two arms within each saved revision, with persistent retrieval and
learning disabled.
The readout study holds questions and checkpoints
fixed while varying the supplied context.
Baselines use their fixed recorded
trajectories.
This panel measures attained information identification and
use on common field references.
These records are separate from the benchmark-level aggregation protocol.

\paragraph{Benchmark result identities.}
The MobileWorld app-count table is a grouped view of the main-result
register, and the Full TAM row in the ablation table uses the same Full
TAM register.
Diagnostic records retain their own source
identities.
Links to the main registers are made by exact episode identity,
not merely by shared task identifiers.
This structure connects the
reported comparisons without conflating historical configurations.

\paragraph{References and units.}
Field judgments combine benchmark definitions, source observations,
executed actions, and destination observations. Audits are model-assisted
and accompanied by their evidence. Unresolved judgments remain separate.
Table~\ref{tab:analysis_cohorts} distinguishes tasks, episodes, operations,
fields, and repeated observations within tasks.

\begin{table*}[t]
\centering\small
\setlength{\tabcolsep}{4pt}
\renewcommand{\arraystretch}{1.10}
\begin{tabularx}{\textwidth}{@{}>{\raggedright\arraybackslash}p{.16\textwidth}c >{\raggedright\arraybackslash}X >{\raggedright\arraybackslash}p{.23\textwidth}@{}}
\toprule
Analysis & Source & Inclusion rule and purpose & Evaluated units \\
\midrule
Diagnostic scope & B & Every official GUI-only task. Task definitions identify the acquired-content transfers to which field scoring applies. & 117 tasks; 52 transfer tasks \\
Information use & H & Common references with a source and a downstream destination. The same fields are scored for all five methods. & 21 tasks; 52 values \\
Value persistence & H & Source-verifiable acquisitions followed at every later decision. The temporal comparison requires ten subsequent decisions to keep the same values throughout. & 202 bindings in 32 tasks; ten-step subset: 106 in 24 tasks \\
Operation preparation & P & Structurally selected diagnostic tasks. All catalogued operations enter the full panel. Shared-source cases test organization when both runs encounter the required inputs. & 20 operation instances; 10 shared-source instances containing 18 fields \\
Readout & Q & Cross-app access to captured facts, with at most two claims per task, then all source-resolvable queries across the same five contexts. & 20 queries; 13 tasks; 100 responses \\
Supervision scope & H & Supported candidates with confirmed persistent writes, paired before and after review; equal task weights. & 335 segments; 215 episodes; 36 tasks \\
Learning material & H & Outcome candidates not nominated by terminal evaluator success. Complete source-value claims with external references support value-level scoring. & 20 outcomes; 372 resource candidates; 234 complete values \\
Partial-result timing & H & Every failed execution with a source-checked correct candidate, providing an observed onset for timing. & 29 executions; 16 tasks; 41 candidates \\
New-input use & H & Actual deliveries define reuse opportunities. First subsequent bindings and traceable text entries define the value and destination tests. & 206 deliveries; 86 binding fields; 29 text-entry fields \\
Computation & H & Every valid archived execution using a shared prompt format, so the measured context block has consistent boundaries. & 250 episodes; 96 tasks; 7,107 actor prompts \\
\bottomrule
\end{tabularx}
\caption{Sources and statistical units of the diagnostic analyses. B: benchmark metadata; H: historical executions; P: paired state control; Q: fixed-checkpoint readout.}
\label{tab:analysis_cohorts}
\end{table*}

\paragraph{Persistence and controlled state execution.}
The fixed archive contains 756 valid full-memory executions across 115
tasks. Inclusion requires a value in its cited read-only UI source and in
the actual actor state. This defines the 32-task acquisition cohort.
Instruction constants and echoes of the agent's own input are excluded.
Typed matching covers email addresses, URLs, dates,
times, phone numbers, and identifiers with at least seven digits.
Other text requires eight alphanumeric characters. Every later prompt enters
the score, including those missing the value. Cross-app decisions yield
2,509 pairs from 24 tasks. Absence from current XML and recent text yields
1,024 pairs from 22 tasks. Each condition includes all matching decisions.
These are literal-presence measurements. Images and other prompt blocks
are outside the immediate-text comparison. The temporal plot instead
requires ten complete subsequent decisions per binding, yielding 106
bindings across 24 tasks. This duration criterion keeps the same values
at every step and uses neither episode success nor subsequent retention.
Longer follow-up requirements define separate cohorts. At 15 decisions,
state and current XML/recent text contain 98.8\% and 34.3\% of tracked
values (88 bindings, 20 tasks). At 20 decisions, the corresponding
figures are 81.5\% and 60.6\% (67 bindings, 18 tasks). These endpoints use
equal task weights within each cohort.

The paired state panel uses Qwen3-VL-32B-Instruct, temperature 0, and a 50-action limit. Both
saved revisions use seed 7. The flat-state intervention includes associated
guidance and output examples. Both arms disable persistent retrieval,
writes, and learning. Section~\ref{app:state_control} extends the main-text
paired runs. Shared source exposure is determined after execution.
Full-panel results accompany this conditional comparison.

\paragraph{Readout selection and inputs.}
The selector scans 117 historical task records. The 53 first occurrences of a stored claim in a later
application yield 32 occurrences with valid source and destination apps
in declared multi-app tasks, or 26 distinct claims across tasks. Seed
20261003 and a cap of two claims per task select 23. Source review resolves
20, including one incorrect stored claim. Three have unresolved source-object,
quantity, or temporal references. Thus the test concerns access to facts
previously captured in TAM state.

All conditions use Qwen3-VL-32B-Instruct at temperature 0. They share the current screenshot, XML,
recent actions, task, and environment policy. An initial probe omitted the
environment policy, including the device date. Corrected inputs were then
applied to every condition with questions, references, and checkpoints
unchanged. Only corrected responses enter this exploratory comparison.
Visual history adds up to seven earlier screenshots. Query-blind notes use
the full prefix and the same image sample, with a 128-word target. Reported
tokens cover readout, excluding memory and note construction. Source-to-query
gaps range from one to 11 steps, with median two.

Omitting each task in turn leaves an 8.3--16.7 percentage-point TAM advantage
over notes and a 0--4.2 point difference from full history. These ranges
measure sensitivity, not confidence intervals. Paired and field-error
analyses reuse all 100 responses. The 28 requested fields are nested within
20 queries. The five multi-field queries contain 13 requested fields.
Error categories describe returned values, not internal causes.
For context $c$, relative error reduction is
$100(E_c-E_{\mathrm{TAM}})/E_c$, where $E$ averages query errors within
tasks and then equally across tasks. All contexts use the same questions
and fixed weights. Error rates are 3.85\% for TAM, 19.23\% for notes,
7.69\% for full history, 11.54\% for visual history, and 73.08\% for recent
context. Relative reductions differ from percentage-point accuracy gains.

\paragraph{Candidate census and source references.}
Of 756 valid executions, 751 match a retrospective review record by namespace
and episode identifier. The 430 executions without candidates remain in
the inventory. The other 321 produce 582 proposals across 48 tasks, including
120 outcomes and 462 resources.

Of 105 admitted outcomes, 85 nominated by native evaluation are excluded
from the independent outcome test. All 20 remaining proposals are included.
Two supported proposals fail source-validity normalization and three
unresolved proposals have malformed review JSON. Among the 15 with usable
review decisions, TAM stores ten supported outcomes and none of the five
contradicted outcomes.

All 372 admitted resource candidates enter Table~\ref{tab:full_source_accounting}.
Official task roles define 297 reference slots in 266 candidates. The
234 complete slots span 210 candidates and 146 executions. The 63 unresolved
slots include missing dates, partial claims, and ambiguous source roles.
Another 106 candidates lack static acquired-value references or concern
instruction-given content. Unassessed items receive no correctness label.
Field judgments assess recorded values, not every statement in an operator.
The local-outcome panel uses all 20 eligible proposals. Unresolved retained
items remain in its correctness denominator, and processing failures do
not remove supported items from coverage. The source-value panel uses
all 234 complete fields across 20 tasks. A field from task $t$ receives
weight $1/n_t$, with $n_t$ fixed from that task's complete fields before
selection. Correctness divides retained supported weight by all retained
weight. Coverage divides retained supported weight by all supported
weight in the pool. Without weighting, TAM has 94.9\% correctness and
78.2\% coverage, versus 98.8\% and 76.4\% for success-only selection.

\begin{table}[!htbp]
\centering\small
\setlength{\tabcolsep}{3pt}
\begin{tabular}{@{}lrrr@{}}
\toprule
Selection & Supported & Contradicted & Unassessed \\
\midrule
All admitted & 189 & 16 & 167 \\
Success-only & 148 & 2 & 87 \\
TAM storage & 150 & 7 & 108 \\
\bottomrule
\end{tabular}
\caption{Accounting for all 372 admitted resource candidates.
A candidate is supported when all its specified reference slots agree,
contradicted when any disagrees, and otherwise unassessed. This
candidate-level table complements the value-level quality comparison.}
\label{tab:full_source_accounting}
\end{table}

\paragraph{Timing and reuse of partial results.}
All 41 reference-correct failed-source candidates have original proposal
indices and complete logs, covering 29 executions across 16 tasks. Each
task receives weight $1/16$, divided equally among its qualifying executions.
Execution progress is the one-based record position divided by the number
of logged decisions. The median is the smallest observed fraction whose
cumulative task weight reaches 50\%. Final-decision records remain included.

Both curves use the same execution weights. The second curve uses the
first correct record that is eventually retained, located at its original
recording position. Executions without retained records contribute zero
throughout. This curve describes the origin of retained material rather
than the time of retrospective storage. Local correctness covers specified
source values, not every claim in a learned operator.

Table~\ref{tab:partial_progress_types} uses equal task weights within each
information type. Material retention averages each task's retained fraction
of correct records. Execution retention and midpoint coverage average
within-task execution fractions. Timing uses each execution's first record
of the relevant type. The overall row deduplicates executions. All eligibility
rules are unchanged from the original source-value audit.

Of the 29 stored failed-source candidates, operators linked to 15 appear
in 56 later episode--operator pairs across 26 tasks. Twenty pairs have
a subsequent new binding, yielding 25 fields with 18 correct, two incorrect,
and five unresolved values. Ten correct destination uses occur in ten
executions. Each target unit is counted once despite multiple source
contributions. This accounting connects checked source values, operator
delivery, and subsequent behavior.

\paragraph{Reuse and computation.}
Chronological reuse is checked within namespaces by matching retrieved
operator identifiers to saved actor prompts. Across all 756 valid runs,
206 first episode--operator deliveries occur in 168 executions across 60 tasks.
They comprise 165 acquisitions, 34 interface outcomes, and seven unresolved
types. The first subsequent binding is fixed before scoring, preventing
later repairs from replacing an incorrect initial value.

Changed-input eligibility uses target references, independently of target
correctness. Comparisons require the same schema, semantic field, and value
type. Compound source records are resolved to the matching field. The
86 target fields include 24 changed inputs, eight unchanged references,
six inputs not newly acquired, and 48 without determinate comparisons.
Destination scoring checks every literal-entry match against subsequent
observations and permits repairs. The 57 unmatched fields receive no
destination judgment. Other GUI input modes lie outside this test.
Delivery and subsequent behavior are observed. Establishing the incremental
benefit of a particular source contribution requires controlled memory-bank
comparisons.

Of 784 valid archived executions, 250 from 96 tasks
share the actor prompt format.
All qualifying episodes contribute within-task averages. Token shares
include repair calls. Character shares cover the state-and-guidance block,
excluding earlier execution recall. The ledger also retains 214 infrastructure-invalid
attempts and covers all 117 task identifiers.

\end{document}